\documentclass[10pt,twocolumn,letterpaper]{article}

\usepackage{wacv}              %

\renewcommand{\paragraph}[1]{%
  \par\vspace{0.25\baselineskip}%
  \noindent\textbf{#1}%
}

\usepackage[accsupp]{axessibility}

\definecolor{wacvblue}{rgb}{0.21,0.49,0.74}
\usepackage[pagebackref,breaklinks,colorlinks,allcolors=wacvblue]{hyperref}

\def\wacvPaperID{233} %
\def\confName{WACV}
\def\confYear{2026}

\title{Mitigating the Modality Gap: Few-Shot Out-of-Distribution Detection with Multi-modal Prototypes and Image Bias Estimation}%

\author{Yimu Wang, Evelien Riddell, Adrian Chow, Sean Sedwards, Krzysztof Czarnecki\\
University of Waterloo\\
{\tt\small \{yimu.wang,evelien.riddell,ahtchow,sean.sedwards,krzysztof.czarnecki\}@uwaterloo.ca}
} 

\usepackage{booktabs}
\usepackage{multirow}
\usepackage{graphicx}
\usepackage{amsmath}
\usepackage{xspace}
\DeclareMathOperator*{\argmax}{arg\,max}

\allowdisplaybreaks
 
\newcommand{\ours}{\textsc{suPreMe}\xspace}
\newcommand{\ITC}{image-text consistency\xspace}
\newcommand{\BPG}{biased prompt generation\xspace}
\usepackage{colortbl}

\usepackage{amsthm}

\newtheorem*{remark}{Remark}
\newtheorem{theorem}{Theorem}
\newtheorem{assumption}{Assumption}

\usepackage{makecell}

\usepackage{amsbsy}
\usepackage{stix}   %

\begin{document}
\maketitle
\begin{abstract}
Existing vision-language model (VLM)-based methods for out-of-distribution (OOD) detection typically rely on similarity scores between input images and in-distribution (ID) text prototypes.
However, the modality gap between image and text often results in high false positive rates, as OOD samples can exhibit high similarity to ID text prototypes.
To mitigate the impact of this modality gap, we propose incorporating ID image prototypes along with ID text prototypes. 
We present theoretical and empirical evidence indicating that this approach enhances VLM-based OOD detection performance without any additional training. 
To further reduce the gap between image and text, we introduce a novel few-shot tuning framework, \ours, comprising \BPG (BPG) and \ITC (ITC) modules. 
BPG enhances image-text fusion and improves generalization (prevents overfitting on the training data) by conditioning ID text prototypes on the Gaussian-based estimated image domain bias;
ITC 
reduces the modality gap by minimizing intra- and inter-modal distances. 
Moreover, inspired by our theoretical and empirical findings, we introduce a novel OOD score $S_{\textit{GMP}}$, leveraging uni- and cross-modal similarities. 
Finally, extensive experiments demonstrate that \ours consistently outperforms existing VLM-based OOD detection methods.
\end{abstract}
    
\section{Introduction}
\label{sec:intro}

\begin{figure}
    \centering
    \scriptsize
    \textsf{
    \begin{tabular}{rclcl}
        Image embeddings: & $\bigcirc$ & \hspace{-1em}ID  & {\small$\bigwhitestar$} & \hspace{-1em}OOD\\
        Prototypes:       & $\mdlgwhtdiamond$ & \hspace{-1em}text & $\square$ & \hspace{-1em}image
    \end{tabular}
    }
    \subcaptionbox{Previous methods}{
        \includegraphics[scale=0.85]{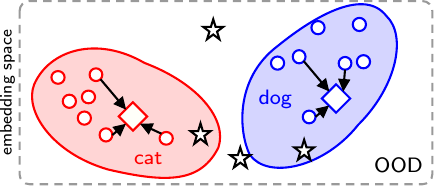}
    }
    \subcaptionbox{Our method: \ours}{
        \includegraphics[scale=0.85]{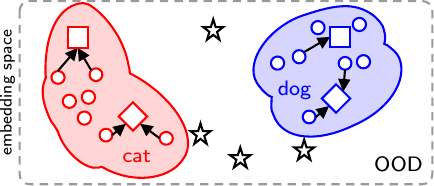}
    }
    \vspace{-1.5em}
    \caption{
    (a) Previous VLM-based OOD detection~\cite{ming2022delving,miyai_locoop_2023,li_learning_2024} uses only ID text prototypes ($\mdlgwhtdiamond$) to identify OOD samples. (b) \ours employs ID image prototypes ($\square$) to complement ID text prototypes, reducing the impact of the image-text modality gap~\cite{liang2022mind} and sharpening the boundary between ID and OOD.}
    \label{fig: teaser}
    \vspace{-1.5em}
\end{figure}

Detecting out-of-distribution (OOD) samples~\cite{hendrycks_baseline_2017,liang_enhancing_2020,liu_energy-based_2020,huang_mos_2021,fort_exploring_2021,Wang_2022_CVPR,sun_out--distribution_2022} is essential for the real-world deployment of machine learning models~\cite{heDeepResidualLearning2016,radford_learning_2021,yang_openood_2022}, as novel samples may emerge and should be flagged for careful consideration. 
Recently, inspired by the power of vision-language foundation models (VLMs)~\cite{radford_learning_2021,girdhar_imagebind_2023,Hess_2024_WACV}, 
novel approaches to OOD detection using VLMs~\cite{fort_exploring_2021,esmaeilpour_zero-shot_2022,ming2022delving,adaloglou_adapting_2023,liu_category-extensible_2023,miyai_locoop_2023,park_powerfulness_2023,wang_clipn_2023,bai_id-like_2024,cao_envisioning_2024,jiang2024negative,li_learning_2024,zhang_lapt_2024,zhang2024adaneg,li2025synood}
have gained significant attention. 
Early VLM-based OOD detection works~\cite{ming2022delving,wang_clipn_2023,jiang2024negative,cao_envisioning_2024} mainly focus on using CLIP~\cite{radford_learning_2021} in a zero-shot setting, where only the in-distribution (ID) class names are utilized. 
For example, Maximum Concept Matching (MCM)~\cite{ming2022delving} 
measures the similarity between input images and text embeddings of ID classes, also referred to as ID \emph{text prototypes}, in the joint vision-language representation space of CLIP.
MCM then uses this similarity score to differentiate OOD and ID samples, on the basis that ID samples should have higher similarity scores. 

Following MCM, other zero-shot methods~\cite{wang_clipn_2023,jiang2024negative,cao_envisioning_2024} aim at generating OOD (negative) text prototypes by querying large language models (LLMs)~\cite{khattab_colbert_2020,openai_gpt-4_2023} or WordNet~\cite{DBLP:journals/cacm/Miller95}. 
These methods obtain OOD scores by evaluating the similarity difference between ID and OOD text prototypes.
To further improve performance, various few-shot tuning OOD detection methods~\cite{liu_category-extensible_2023,li_learning_2024,zhang_lapt_2024,bai_id-like_2024,nie_out--distribution_2024} have been proposed. 
These methods mainly target learning OOD (negative) text prototypes. 
For example, NegPrompt~\cite{li_learning_2024} learns OOD text prototypes by minimizing the similarity between OOD text prototypes and ID data with contrastive learning.

\begin{figure}
    \centering
    \includegraphics[width=0.9\linewidth]{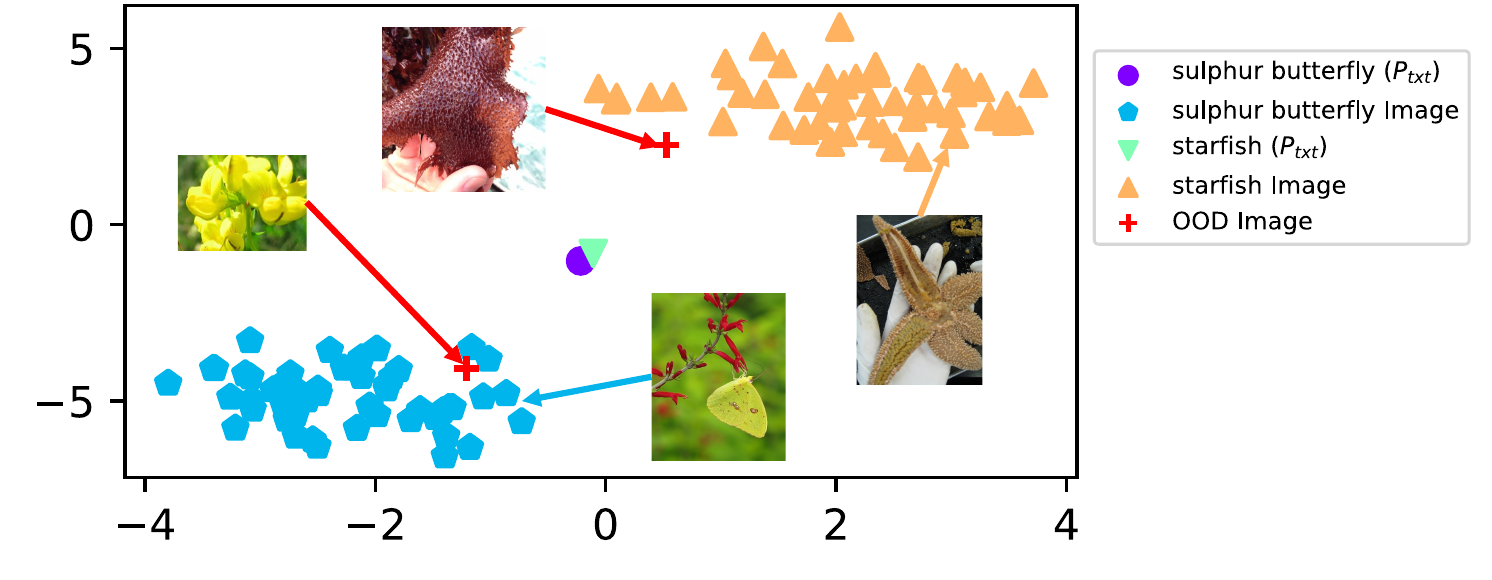}
    \vspace{-1.5em}
    \caption{
    Examples where OOD images (iNaturalist) are located closer to ID text prototypes compared to ID images (ImageNet-1k) generated by CLIP (ViT-B/16). We also visualize the embeddings generated by our \ours in \cref{fig: tsne viz}, which has smaller gaps between image and text embeddings.}
    \label{fig: ood example}
    \vspace{-1.3em}
\end{figure}

While these approaches have shown promise using CLIP, they use only text prototypes and thus are limited by the modality gap~\cite{liang2022mind} between image and text modalities. 
Liang \etal~\cite{liang2022mind} demonstrate that CLIP’s image and text embeddings are located in two clearly separate regions of the embedding space.
As a result, OOD images may also exhibit high similarity to ID text prototypes 
(\cref{fig: ood example}, more examples are in \cref{ablations: fig: ood example}),
since high similarity can stem from either true semantic similarity or mere spatial closeness to the ID text prototypes, causing a risk of increased false positives. 
This observation raises a critical question about the impact of the modality gap on VLM-based OOD detection, leading to our first research question: 
    \vspace{-0.45em}
    \begin{center}
        \textit{RQ1: What is the impact of the modality gap on VLM-based OOD detection methods?}
    \end{center}
    \vspace{-0.45em}

In response to RQ1, we hypothesize that the modality gap negatively impacts performance and that incorporating image and text (multi-modal) prototypes, as opposed to only text prototypes, could alleviate this effect. We illustrate this in~\cref{fig: teaser}.
To explore this, we compute class-specific ID image prototypes by averaging embeddings extracted from the ID data 
within each class. 
We extend the MCM score $S_{\textit{MCM}}$~\cite{ming2022delving} to incorporate the image prototypes alongside the text prototypes. 
Compared to $S_{\textit{MCM}}$, our empirical analysis shows that, without any training, using the extension of $S_{\textit{MCM}}$ with multi-modal prototypes improves average FPR95 and AUROC from 32.4 and 94.5 to 24.2 and 95.8, respectively, on Imagenet-100~\cite{dengImageNetLargescaleHierarchical2009} (ID) across four OOD datasets (\cref{tab: MMA}).
For theoretical support, in \cref{the: image anchor helps}, we demonstrate that incorporating image prototypes increases the score separation between ID and OOD data, leading to improved performance. 

Given these findings, \ie, that the modality gap significantly affects VLM-based OOD detection, we conclude that it is necessary to explore approaches mitigating the gap to improve reliability. 
This leads us to our second question:
\vspace{-0.45em}
\begin{center} 
    \textit{RQ2: How can we mitigate the impact of modality gap to improve VLM-based OOD detection?} 
\end{center}
\vspace{-0.45em}
To address RQ2, we propose a novel few-\textbf{\textsc{s}}hot tuning, m\textbf{\textsc{u}}lti-modal \textbf{\textsc{pr}}ototyp\textbf{\textsc{e}}-based \textbf{\textsc{me}}thod for OOD detection with CLIP, termed \ours. 
\ours comprises \BPG (BPG) and \ITC (ITC) modules. 
BPG introduces a Gaussian-based image domain bias to enhance image-text fusion and generalization on ID data unseen during fine-tuning by preventing overfitting on the limited training ID data. 
ITC reduces the modality gap by minimizing inter- and intra-modal distances as CLIP remains frozen. 
\ours also makes use of a new OOD score, which we call the \textbf{G}eneralized \textbf{M}ulti-modal \textbf{P}rototype OOD score, denoted $S_{\textit{GMP}}$. 

\noindent
\textbf{BPG}. 
In contrast to previous OOD detection methods that only utilize learnable contexts for text prototypes~\cite{li_learning_2024,miyai_locoop_2023,bai_id-like_2024}, BPG conditions text prototypes on three components: learnable contexts, the image embedding, and the ID image domain bias. 
The image embedding facilitates image-text fusion, enhancing cross-modal alignment. 
The third component, the Gaussian-based ID image domain bias, %
captures the distribution of ID image embeddings to improve generalization on unseen ID data and avoid overfitting.

\noindent
\textbf{ITC}. 
As the CLIP remains frozen during training, to minimize the modality gap, we introduce intra- and inter- modal losses with two mappings.
First, the image embedding $I$ is mapped to the text domain as $I' = f_\textit{img-txt}(I)$ with the image-to-text mapping $f_\textit{img-txt}$.
To ensure it aligns with ID text prototypes, we employ the inter-modal loss $\ell_{\textit{inter}}$. 
Additionally, to avoid information loss, the intra-modal loss $\ell_{\textit{intra}}$ is applied between the original embedding $I$ and the reconstructed image embedding $\hat{I}=f_\textit{txt-img}(I')$ using the text-to-image mapping $f_\textit{txt-img}$.

\noindent
$\pmb{S_{\textit{GMP}}}$. 
Building on our findings%
, we introduce a new OOD score, %
$S_{\textit{GMP}}$, which integrates uni- and cross-modal similarities. 
While previous methods rely solely on the similarity between ID text prototypes and input image embedding, $S_{\textit{GMP}}$ incorporates the similarity between multi-modal input embeddings---the image embedding $I$ and the mapped image embedding $f_{\text{img-txt}}(I)$---and ID multi-modal (image and text) prototypes, respectively.

Our contributions are summarized as follows:
\begin{itemize}
    \item 
        (RQ1) 
        We empirically and theoretically demonstrate that multi-modal (image and text) prototypes reduce the negative impact of the modality gap, resulting in performance improvements without additional training.

    \item 
        (RQ2) To further mitigate the modality gap, we propose a novel few-shot tuning framework, \ours, comprising \BPG (BPG) and \ITC (ITC) modules.
        
    \item 
        (RQ2) Building on our empirical and theoretical results, we design a new OOD score, $S_{\textit{GMP}}$, exploiting ID multi-modal prototypes and multi-modal input embeddings to enhance performance and robustness. 
        
    \item
        (RQ2) %
        Extensive experiments across multiple benchmarks~\cite{dengImageNetLargescaleHierarchical2009,zhang_openood_2023}
        demonstrate that \ours outperforms existing OOD detection methods.
\end{itemize}

\section{Related Work}

\textbf{Out-of-distribution (OOD) detection.}
OOD detection~\cite{hendrycks_baseline_2017,huang_importance_2021,liu_energy-based_2020,yang_openood_2022,zhang_openood_2023} aims to discriminate ID samples and OOD samples. 
Traditional methods include post-hoc~\cite{sun_react_2021,lee_simple_2018,park_nearest_2023,sun_out--distribution_2022}, 
confidence enhancement~\cite{devries_learning_2018,hein_why_2019,wei_mitigating_2022,xu_vra_2023,narasimhan_plugin_2023}, and outlier exposure methods~\cite{hendrycks_deep_2018,jiang_dos_2023,du_dream_2023,kieuOutlierDetectionMultidimensional2018,huang_mos_2021,liu_residual_2023,zhou_learning_2021,fort_exploring_2021}.
Recently, the field has shifted towards exploiting large-scale pre-trained VLMs~\cite{radford_learning_2021,girdhar_imagebind_2023,Hess_2024_WACV}, \ie, CLIP~\cite{radford_learning_2021}, to enhance OOD detection.
MCM~\cite{ming2022delving}
employs CLIP and explores the effects of softmax and temperature scaling in a zero-shot manner. 
Subsequent works can be roughly categorized into zero-shot and few-shot tuning methods. 
Zero-shot methods~\cite{ming2022delving,jiang2024negative,cao_envisioning_2024,wang_clipn_2023} primarily focus on identifying negative classes by querying LLMs~\cite{khattab_colbert_2020,openai_gpt-4_2023} or WordNet~\cite{DBLP:journals/cacm/Miller95}. 
In parallel, few-shot tuning methods~\cite{liu_category-extensible_2023,li_learning_2024,zhang_lapt_2024,bai_id-like_2024,nie_out--distribution_2024} also mainly target finding negative text prompts (OOD text prototypes). 
For example, CATEX~\cite{liu_category-extensible_2023}, LSN~\cite{nie_out--distribution_2024}, and NegPrompt~\cite{li_learning_2024} learn negative prompts by minimizing the similarity between negative prompts and ID training data. 
Following those works, SynOOD~\cite{li2025synood} generates OOD images using image generative models and LLMs to finetune the image and text encoders. 
On the other side, AdaNeg~\cite{zhang2024adaneg} introduces a memory bank during testing to selectively cache discriminative features from test images, representing the targeted OOD distribution. 
The memory bank further helps in distinguishing new OOD images.

In contrast to previous VLM-based methods that use only text prototypes, \ours enhances OOD detection by employing multi-modal prototypes and encouraging cross-modal alignment to reduce the modality gap.

\noindent
\textbf{Prompt tuning}. 
Prompt tuning originates in Natural Language Processing~\cite{lester_power_2021,li_prefix-tuning_2021,shin_autoprompt_2020} as a method for automating template/prompt creation in models such as BERT~\cite{devlinBERTPretrainingDeep2019} and GPT~\cite{openai_gpt-4_2023}. 
For example, AutoPrompt~\cite{shin_autoprompt_2020} is a gradient-based approach for identifying ``optimal'' prompts, replacing manually designed prompts. 
Recently, prompt tuning has been applied in computer vision models~\cite{khattak_self-regulating_2023,zhou2022cocoop,zhou_learning_2022}.
Notably, CoOp~\cite{zhou_learning_2022} and CoCoOp~\cite{zhou2022cocoop}, as representative methods in visual prompt tuning (VPT), employ learnable prompts optimized by minimizing classification loss to improve CLIP's performance. 

Different from previous VPT works~\cite{zhou2022cocoop,zhou_learning_2022,miyai_locoop_2023,liu_category-extensible_2023,bai_id-like_2024,nie_out--distribution_2024} that generate text prompts with text and individual image data, \ours generates text prompts/prototypes with a novel Gaussian-based image domain bias for estimating the distribution of image data, which enables enhanced image-text fusion with improved generalization ability.

\section{Preliminaries}

\subsection{Problem Setup}

\begin{figure*}
    \centering
    \includegraphics[width=\linewidth]{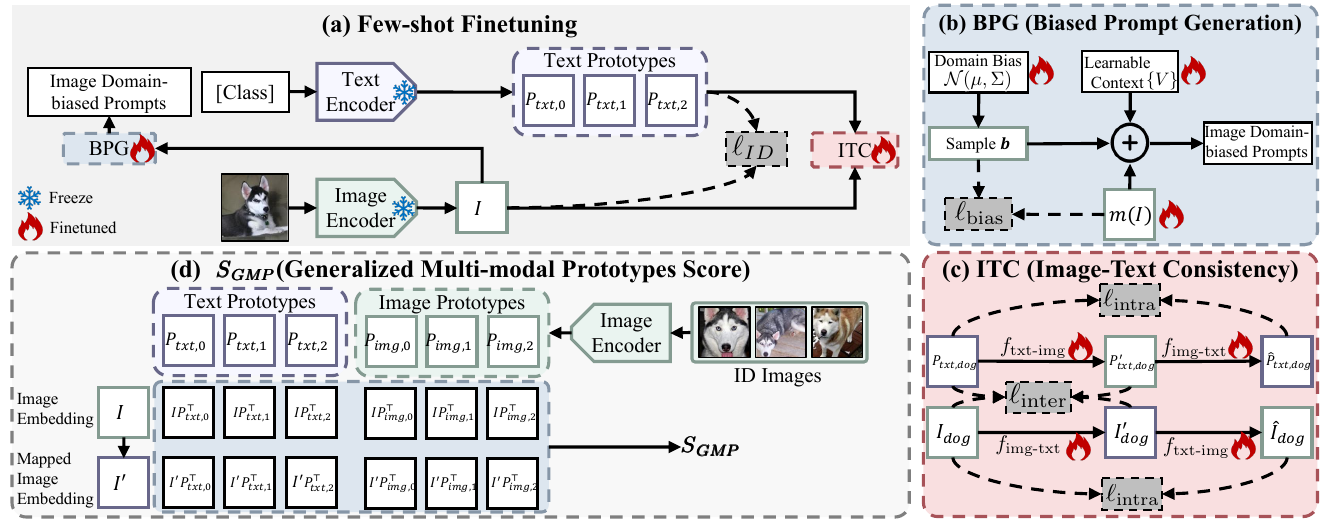}
    \vspace{-2em}
    \caption{
    Overview of \ours. 
    The two novel modules, \ie, BPG (\BPG) and ITC (\ITC), are designed to minimize the modality gap. 
    (a) During the few-shot fine-tuning stage, (b) BPG generates image domain-biased prompts, conditioned on the estimated image domain bias and the mapped image embedding $m(I)$ for better image-text fusion.
    (c) ITC minimizes the modality gap directly by the intra- and inter-modal losses ($\ell_{\textit{intra}}$ and $\ell_{\textit{inter}}$) with the text-to-image mapping $f_\textit{txt-img}$ and the image-to-text mapping $f_\textit{img-txt}$. 
    (d) During inference, image prototypes are obtained by averaging each class's base ID image embeddings (\cref{eq: image anchor}). 
    The proposed $S_{\textit{GMP}}$ (\cref{eq: MMO}) is calculated based on the maximum similarity between the multi-modal embeddings (the image embedding $I$ and mapped image embedding $I'$) and ID multi-modal prototypes (\ie, text prototypes $\{P_{\textit{txt},c}\}_{c=1}^{C}$ and image prototypes $\{P_{\textit{img},c}\}_{c=1}^{C}$). 
    $S_{\textit{MCM}}$ refers to the MCM score~\cite{ming2022delving}.
    }
    \label{fig: main figure}
    \vspace{-1em}
\end{figure*}

Let and $\mathcal{Y}=\{y_{1}, \ldots, y_{C}\}$ represent the set of ID class labels, where $C$ is the number of ID classes. 

\noindent \textbf{OOD detection}. 
In real-world applications, AI models are trained on ID data and may misclassify OOD data into ID classes with high confidence~\cite{joseph_towards_2021}. 
To tackle this problem, OOD detection~\cite{yang_generalized_2024} is proposed to identify OOD samples using a score function $S(\cdot)$~\cite{li_rethinking_2023,li_moodv2_2024}. 
A sample is classified as OOD if $S(\mathbf{x}) \leq \gamma$, where $\gamma$ is a predefined threshold. 

\noindent \textbf{Few-shot tuning}. 
In contrast to existing works that either utilize the entire ID training dataset~\cite{li_tagood_2024,Wang_2022_CVPR,lee_simple_2018} or avoid tuning entirely~\cite{wang_clipn_2023,ming2022delving,cao_envisioning_2024,jiang2024negative}, we study the scenario where the model is fine-tuned using only a subset of the ID training data (16 images per class) {without} access to OOD data or other additional data. 

\subsection{Revisiting CLIP, MCM, and Prompt Tuning}

\textbf{CLIP~\cite{radford_learning_2021}} is a foundational VLM pre-trained on a web-scale image-text dataset using self-supervised contrastive learning~\cite{DBLP:conf/icml/ChenK0H20}.
Specifically, for classification, CLIP first incorporates class labels $\mathcal{Y} = \{y_{c}\}_{c=1}^{C}$, \eg, ``cat'', into fixed pre-designed, rather than learned, text prompts, \eg, ``\texttt{a photo of a {\it label\,}}''.
These prompts combined with class labels are processed by CLIP’s text encoder $f_{\textit{text}}(\cdot)$ to generate text embeddings $\{P_{\textit{txt},c}\}_{c=1}^{C}$, where $P_{\textit{txt},c} = f_{\textit{text}}(\text{``}\texttt{a photo of a }\{y_c\}\text{''})$.
Given an image $\mathbf{x}$, the image embedding $I$ is obtained by the image encoder $f_{\textit{image}}(\cdot)$ as $I = f_{\textit{image}}(\mathbf{x})$. 
The index of the predicted class label is given by $\argmax_{c \in \{1..C\}} \cos(I, P_{\textit{txt},c})$, where $\cos(\cdot, \cdot)$ denotes the cosine similarity.

\noindent
\textbf{CLIP for OOD Detection (MCM)~\cite{ming2022delving}}.
Beyond its strong classification capabilities, MCM demonstrates that pre-trained CLIP models also exhibit robust zero-shot OOD detection capabilities. 
Specifically, MCM defines the maximum classification score as the OOD score $S_{\textit{MCM}}$,
\begin{multline}
\label{eq: MCM}
    S_{\textit{MCM}} (I ,\{P_{\textit{txt},c}\}_{c=1}^{C}) = \\\max_{c\in \{1..C\}} \frac{\exp(\cos(I, P_{\textit{txt},c}) / \tau)}{\sum_{j=1}^{C} \exp(\cos(I, P_{\textit{txt},j}) / \tau)},
\end{multline}
where $\tau$ is the temperature and $\exp(\cdot)$ is the exponential function.

\noindent
\textbf{Visual prompt tuning~\cite{zhou_learning_2022,zhou2022cocoop,guo_pfedprompt_2023,yao_visual-language_2023}}. 
To improve CLIP's performance when target training data is accessible, %
CoOp~\cite{zhou_learning_2022} replaces manually designed prompt templates with learnable soft prompts (context tokens) $\textit{CoOp}_c=[V_1, \ldots, V_L, y_c]$ for class index $c$. $L$ is the length of learnable prompts and each $V_i$ represents a learnable vector. 
The class-wise text embedding is then generated as $P^{\textit{CoOp}}_{\textit{txt}, c}=f_{\textit{text}}(\textit{CoOp}_c)$ for class index $c$. 
With CLIP's text and image encoder parameters frozen, the learnable prompts are optimized using the cross-entropy loss:
\begin{equation}
    \ell_{\textit{ID}}(I) = -\log \frac{\exp({\cos(I, P_{\textit{txt}, c}^{\textit{CoOp}} })/ \tau)}{\sum_{j=1}^{C} \exp({\cos(I, P_{\textit{txt}, j}^{\textit{CoOp}}})/\tau)}\,,
\end{equation}
where $c$ is the index of the ground truth label for the input image. 
Following CoOp, to avoid potential overfitting on the training data, CoCoOp~\cite{zhou2022cocoop} 
conditions prompts on the image embedding $I$ as $\textit{CoCoOp}_c=[V_1 + m(I),\ldots, V_L+m(I), y_c]$, 
where $m(\cdot)$ is a two-layer multi-layer perception (MLP, Linear-ReLU-Linear).

\section{Methodology}

In this section, we first answer RQ1 by presenting our hypothesis and findings in \cref{Sec: MMA}, where we demonstrate that the modality gap negatively impacts performance and that incorporating image prototypes alleviates this effect.
Then, to mitigate the modality gap (RQ2), we introduce \ours in \cref{sec: our method} with two modules and a novel OOD score, $S_{\textit{GMP}}$. We provide an overview of \ours in \cref{fig: main figure}. 

\subsection{RQ1: Multi-modal Prototypes}
\label{Sec: MMA}

CLIP has been applied in different areas~\cite{tschannen_clippo_2023,gao_clip2tv_2022}. 
However, a recent study~\cite{liang2022mind} demonstrates that representations of images and texts are clearly separated, creating what is known as the modality gap. 
It remains unclear, however, whether this gap positively or negatively impacts OOD detection.

We hypothesize that the modality gap negatively impacts OOD detection performance
and that it can lead to increased false positives, as OOD images may exhibit high similarity to ID text prototypes due to either semantic similarity or small spatial distance. 

To test our hypothesis, we demonstrate that incorporating multi-modal (image and text) prototypes can mitigate the negative impact of the modality gap and reduce false positives. 
We obtain ID text prototypes $\{P_{\textit{txt},c}\}_{c=1}^{C}$ as in MCM~\cite{ming2022delving}. 
For ID image prototypes $\{P_{\textit{img},c}\}_{c=1}^{C}$, we collect ID base images $\{\mathbf{x}_i\}_{i=1}^{N_{\textit{base}}}$ with corresponding labels $\{y_{i}\}_{i=1}^{N_{\textit{base}}}$, where $N_{\textit{base}}$ is the number of base images. 
ID image prototypes are then calculated by averaging the image embeddings for each class label $y_c$:
\begin{equation}
\small
    \label{eq: image anchor}
    P_{\textit{img},c} = \frac{\sum_{i=1}^{N_{\textit{base}}} \mathbb{I}(y_{i}=y_c) f_{\textit{image}}(\mathbf{x}_{i})}{\sum_{j=1}^{N_{\textit{base}}} \mathbb{I}(y_{j}=y_c)} ,
\end{equation}
where $\mathbb{I}(\cdot)$ is the indicator function and $f_{\textit{image}}(\cdot)$ is CLIP's image encoder. 
With these image prototypes, we extend $S_{\textit{MCM}}$ (\cref{eq: MCM})~\cite{ming2022delving} to a new OOD score, which we term the Multi-Modal Prototypes score, $S_{\textit{MMP}}$, defined as
\begin{equation}\label{eq: MMA}
\small
    S_{\textit{MMP}} %
    = \frac{S_{\textit{MCM}} (I, \{P_{\textit{img},c}\}_{c=1}^{C}) + S_{\textit{MCM}} (I ,\{P_{\textit{txt},c}\}_{c=1}^{C})}{2} \,.
\end{equation}
This approach balances contributions from both modalities, potentially mitigating the impact of the modality gap.

\begin{table}[t!]
\centering
\begin{tabular}{l|cc}
\toprule
Methods    & FPR95 (\%) $\downarrow$   & AUROC (\%) $\uparrow$  \\
\midrule
\rowcolor[gray]{.9} \multicolumn{3}{c}{ImageNet-100 $\rightarrow$ 4 Datasets (Average)} \\
$S_{\textit{MCM}}$~\cite{ming2022delving}        & 32.58                & 94.48             \\
\rowcolor{green!10} $S_{\textit{MMP}}$ (Ours, \cref{eq: MMA})        & \textbf{24.18}                & \textbf{95.79}            \\
\bottomrule
\end{tabular}%
\vspace{-1em}
\caption{
Comparison of OOD detection results on ImageNet-100 (ID) $\rightarrow$ 4 OOD datasets (OOD) with CLIP-B/16.
FPR95 represents the false positive rate of OOD images when the true positive rate of ID images is at 95\%. 
AUROC is the area under the
receiver operating characteristic curve.
Best in \textbf{bold}. 
}\label{tab: MMA}
\vspace{-0.6em}
\end{table}

\noindent
\textbf{Empirical evidence}. 
We begin by empirically validating the effectiveness of image prototypes and $S_{\textit{MMP}}$. 
We use ImageNet-100~\cite{dengImageNetLargescaleHierarchical2009} as the ID dataset, with four OOD datasets:  iNaturalist~\cite{van_horn_inaturalist_2018}, SUN~\cite{xiao_sun_2010}, Places~\cite{zhou_places_2018}, and Texture~\cite{cimpoi_describing_2014}. 
With all training images of ImageNet-100 used for obtaining image prototypes, we observe in \cref{tab: MMA} that, on average, $S_{\textit{MMP}}$ outperforms $S_{\textit{MCM}}$ in FPR95 and AUROC, demonstrating the effectiveness of image prototypes and $S_{\textit{MMP}}$.

\noindent
\textbf{Theoretical observation.} 
To further understand why image prototypes improve performance, we present a theoretical analysis in \cref{the: image anchor helps} of the Appendix. 
The theorem shows that $S_{\textit{MMP}}$ increases the expected score separation between ID and OOD samples compared to using text prototypes alone (as in $S_{\textit{MCM}}$) under general assumptions on OOD and ID distributions used by previous works~\cite{fu_clipscope_2025,liang2022mind}. 

While incorporating image prototypes improves performance
, inspired by recent research~\cite{miyai_locoop_2023,li_learning_2024}, we further fine-tune CLIP 
in a few-shot tuning manner to mitigate the impact of the modality gap (RQ2). 

\subsection{RQ2: $\pmb{\text{\ours}}$}\label{sec: our method}

To alleviate the negative impact of the image-text modality gap~\cite{liang2022mind} in CLIP (RQ2), we propose \ours, which is summarized in \cref{fig: main figure}. 
\ours incorporates two modules, \ie, the {\BPG} module (BPG) and the {\ITC} module (ITC), along with a new OOD score. 
BPG enhances image-text fusion using image domain-biased prompts, while ITC minimizes both intra- and inter-modal distances. 
\ours's novel OOD score, $S_{\textit{GMP}}$, exploits uni- and cross-modal similarities to improve robustness and performance. 

\noindent
\paragraph{Biased Prompt Generation (BPG).}
Drawing inspiration from CoCoOp~\cite{zhou_learning_2022}, we employ a set of learnable context tokens $\{V_i \in\mathcal{R}^{1\times N_{\text{lm}}}\}_{i=1}^{L}$ and the mapped image embedding $m(I) \in\mathcal{R}^{1\times N_{\text{lm}}}$ to generate the text prototypes, where $L$ is the length of learnable contexts, $m(\cdot)$ is a two-layer MLP, and $N_{\text{lm}}$ is the dimension of text encoder's input.
Similar to previous methods~\cite{zhou2022cocoop,miyai_locoop_2023}, learnable context tokens replace the manually designed prompt templates, \eg, ``\texttt{a photo of a {\it class\,}}'', for generating text prototypes.

To improve generalization on unseen ID data and avoid overfitting on few-shot training data, we introduce a third component, a Gaussian-based estimated image domain bias $\mathcal{N}(\mathbf{\mu}, \Sigma)$. 
This bias counteracts the modality gap by pre-biasing text prototypes toward ID images and augments the training data to better approximate the true distribution of mapped ID image embeddings. 
Using a Gaussian distribution is a reasonable a priori choice and is mathematically tractable. Our ablation study (\cref{tab: ab: modules}) confirms that this choice is valid and is successful in reducing FPs.
After that, we generate image domain-biased prompts (IDBP) conditioned on these three components. 

\textbf{Estimating image domain bias}. 
In each training iteration, we sample $\mathbf{b} \in\mathcal{R}^{1\times N_{\text{lm}}}$ from $\mathcal{N}(\mathbf{\mu}, \Sigma)$ as
\begin{equation}
    \small
    \mathbf{b} = \mu + \sigma n, n\sim \mathcal{N}(0, \mathcal{I}),
\end{equation}
where $\Sigma = \sigma \sigma^\top$ (based on the Cholesky decomposition) and $\mathcal{N}(0, \mathcal{I})$ represents the standard Gaussian distribution, given identity matrix $\mathcal{I}$. 
To align the domain bias with the distribution of training ID image embeddings, we employ the following loss:
\begin{equation}
\small
\label{loss: bias}
    \ell_{\textit{bias}}(I) = \| \mu - m(I)\|_1 + \|\mathbf{b} - m(I)\|_1\,,
\end{equation}
where $\|\cdot\|_1$ is the $\ell_1$ distance. 
In the first term of the loss (\cref{loss: bias}), $\mu$ is updated. 
In the second term, as $\mathbf{b}$ is sampled using $\mu$ and $\Sigma$, $\mu$ and $\Sigma$ will be updated during the few-shot training procedure.
The first term, $||\mu - m(I)||_1$, pushes the bias distribution's mean $\mu$ toward the centroid of training image embeddings, while the second term, $||{\mathbf{b}} - m(I)||_1$, ensures sampled bias vectors $\mathbf{b}$ are within the training image embedding space. 

\begin{table*}[t!]
\centering
\resizebox{\textwidth}{!}{%
\begin{tabular}{@{}llc|cccccccc|cc@{}}
\toprule
\multirow{3}{*}{Methods} & \multirow{3}{*}{Venue} & \multirow{3}{*}{Shots} & \multicolumn{8}{c|}{ImageNet-1k $\rightarrow$ OOD Datasets} & \multicolumn{2}{c}{\multirow{2}{*}{Average}} \\
 & & & \multicolumn{2}{c}{iNaturalist} & \multicolumn{2}{c}{SUN} & \multicolumn{2}{c}{Places} & \multicolumn{2}{c|}{Texture} &  \\ \cmidrule(lr){4-5} \cmidrule(lr){6-7} \cmidrule(lr){8-9} \cmidrule(lr){10-11}
 &  & & FPR95 $\downarrow$ & AUROC $\uparrow$ & FPR95 $\downarrow$ & AUROC $\uparrow$ & FPR95 $\downarrow$ & AUROC $\uparrow$ & FPR95 $\downarrow$ & AUROC $\uparrow$ & FPR95 $\downarrow$ & AUROC $\uparrow$ \\
\midrule
MOS$\dagger$~\cite{huang_mos_2021} (BiT) & CVPR'21  & - & 9.28 & 98.15 & 40.63 & 92.01 & 49.54 & 89.06 & 60.43 & 81.23 & 39.97 & 90.11 \\
Fort \etal{}$\dagger$~\cite{fort_exploring_2021} (ViT-B) & NeurIPS'21 & - & 15.07 & 96.64 & 54.12 & 86.37 & 57.99 & 85.24 & 53.32 & 84.77 & 45.12 & 88.25 \\
Energy$\dagger$~\cite{liu_energy-based_2020} (ViT-B)& NeurIPS'20 & - & 21.59 & 95.99 & 34.28 & 93.15 & 36.64 & 91.82 & 51.18 & 88.09 & 35.92 & 92.26 \\
MSP$\dagger$~\cite{hendrycks_baseline_2017} (ViT-B) & ICLR'17 & - & 40.89 & 88.63 & 65.81 & 81.24 & 67.90 & 80.14 & 64.96 & 78.16 & 59.89 & 82.04 \\
ODIN$\ddagger$~\cite{liang_enhancing_2020} & ICLR'18 & - & 30.22 & 94.65 & 54.04 & 87.17 & 55.06 & 85.54 & 51.67 & 87.85 & 47.75 & 88.80\\
ViM$\ddagger$~\cite{Wang_2022_CVPR} & CVPR'22 & - & 32.19 & 93.16 & 54.01 & 87.19 & 60.67 & 83.75 & 53.94 & 87.18 & 50.20 & 87.82\\
KNN$\ddagger$~\cite{sun_out--distribution_2022} & ICML'22 & - & 29.17 & 94.52 & 35.62 & 92.67 & 39.61 & 91.02 & 64.35 & 85.67 & 42.19 & 90.97\\
\rowcolor[gray]{.9} \multicolumn{13}{c}{VLM-based OOD Detection (CLIP of VIT-B/16)}  \\
MCM~\cite{ming2022delving} & NeurIPS'22  & 0 & 30.91 & 94.61 & 37.59 & 92.57 & 44.69 & 89.77 & 57.77 & 86.11 & 42.74 & 90.77 \\
CLIPN~\cite{wang_clipn_2023} & CVPR'23 & 0 & 23.94 & 95.27 & 26.17 & 93.93 & 33.45 & 92.28 & 40.83 & 90.93 & 31.10 & 93.10 \\
EOE~\cite{cao_envisioning_2024} & ICML'24 & 0 &12.29 & 97.52 & 20.40 & 95.73 & 30.16 & 92.95 & 57.53 & 85.64 & 30.09 & 92.96 \\
CoOp*~\cite{zhou_learning_2022} & IJCV & 16& 29.47 & 94.89 & 31.34 & 93.36 & 40.28 & 90.07 & 54.25 & 87.58 & 38.83 & 91.47\\
CoCoOp*~\cite{zhou2022cocoop} & CVPR'22 & 16& 30.74 & 94.73 & 31.18 & 93.15 & 38.75 & 90.63 & 53.84 & 87.92 & 38.63 & 91.61 \\
NPOS~\cite{tao_non-parametric_2023} & ICLR'23 & - & 16.58 & 96.19 & 43.77 & 90.44 & 45.27 & 89.44 & 46.12 & 88.80 & 37.93 & 91.22 \\
LoCoOp~\cite{miyai_locoop_2023} & NeurIPS'23 & 16 & 16.05 & {96.86} & 23.44 & 95.07 & 32.87 & 91.98 & 42.28 & 90.19 & 28.66 & 93.52 \\
CATEX~\cite{liu_category-extensible_2023} & NeurIPS'23 & - & 10.18 & 97.88 & 33.87 & 92.83 & 41.43 & 90.48 & {33.17} & {92.73} & 29.66 & 93.48 \\
LSN~\cite{nie_out--distribution_2024} & ICLR'24 & - & 21.56 & 95.83 & 26.32 & 94.35 & 34.48 & 91.25 & 38.54 & 90.42 & 30.22 & 92.96 \\
ID-Like~\cite{bai_id-like_2024} & CVPR'24 & 4 & 8.98 & 98.19 & 42.03 & 91.64 & 44.00 & 90.57 & \textbf{25.27} & \underline{94.32} & 30.07 & 93.68 \\
NegPrompt~\cite{li_learning_2024} & CVPR'24  & 16 & \underline{6.32} & \textbf{98.73} & 22.89 & 95.55 & {27.60} & \underline{93.34} & {35.21} & {91.60} & 23.01 & {94.81} \\
SCT~\cite{yu_self-calibrated_2024} & NeurIPS'24 & 16 & 13.94 & 95.86 & 20.55 & 95.33 & 29.86 & 92.24 & 41.51 & 89.06 & 26 47 & 93.37 \\
Local-Prompt~\cite{zeng_local-prompt_2025} & ICLR'25 & 16 & 8.63 & 98.07 & 23.23 & 95.12 & 31.74 & 92.42 & 34.50 & 92.29 & 24.52 & 94.48 \\
\midrule
\rowcolor{green!10} \ours & & 16 & \textbf{5.91} & \textbf{98.73} & \textbf{15.85} & \textbf{96.08} & {\textbf{26.18}} & 93.19 & 34.88 & 91.93 & \underline{20.70} & \underline{95.38} \\
\rowcolor{green!10} \ours{}$^{\dagger}$ &  & 16 & 8.27 & \underline{98.29} & \underline{19.40} & \underline{95.84} & \underline{26.69} & \textbf{93.56} & \underline{26.77} & \textbf{94.45} & \textbf{20.28} & \textbf{95.54} \\
\bottomrule
\end{tabular}%
}
\vspace{-0.8em}
\caption{OOD detection results with ID data of ImageNet-1k and four OOD datasets using CLIP (VIT-B/16). Best in \textbf{bold}. Second best is \underline{underlined}.
``*'' is cited from LSN~\cite{nie_out--distribution_2024}. 
``$\dagger$'' represents results from MCM~\cite{ming2022delving}.
``$\ddagger$'' represents results from NPOS~\cite{tao_non-parametric_2023}. 
\ours{}$^{\dagger}$ employs 16 images per class for training and the whole training split for calculating ID image prototypes.
}
\label{tab: main}
\end{table*}

\textbf{Generating image domain-biased prompts (IDBP)}. 
We condition IDBPs on three components: the learnable contexts $\{V_{i} \in\mathcal{R}^{1\times N_{\text{lm}}}\}_{i=1}^{L}$, mapped image embedding $m(I) \in\mathcal{R}^{1\times N_{\text{lm}}}$, and a sample $\mathbf{b} \in\mathcal{R}^{1\times N_{\text{lm}}}$ from $\mathcal{N}(\mathbf{\mu}, \Sigma)$ as
\begin{equation}
    \small
    \textit{IDBP}_c = [V_1 + m(I) + \mathbf{b}, \ldots, V_L + m(I) + \mathbf{b}, y_c]\,.
\end{equation}
The final ID text prototype $P_{\textit{txt}, c}$ for the $c$-th class is obtained as $P_{\textit{txt}, c} = f_{\textit{text}} (\textit{IDBP}_c)$.

We thus establish a foundation for cross-modal fusion by conditioning prompts on learned contexts, image embeddings, and the estimated image domain bias. 

\paragraph{Image-Text Consistency (ITC).}
While BPG reduces the gap by image-text fusion, image-text consistency (ITC) directly reduces the modality gap by aligning image and text embeddings through inter- and intra-modal distances at a fine-grained level.
It consists of two mappings, \ie, the image-to-text mapping $f_\textit{img-txt}(\cdot)$ and text-to-image mapping $f_\textit{txt-img}(\cdot)$, which are used for mapping embeddings into the other modality. 

The first component of ITC is the \emph{inter-modal loss}. 
It is designed to ensure that mapped text/image embeddings ($\{f_\textit{txt-img}(P_{\textit{txt},c})\}_{c=1}^{C}$ and $f_\textit{img-txt}(I)$) remain close to their original counterparts, thus enabling effective alignment. 
It is designed to minimize the cross-entropy loss:
\begin{equation}
\label{loss: inter}
\small
\begin{aligned}
    \ell_{\textit{inter}}(I) = & -\log \frac{\exp({\cos(I, f_\textit{txt-img}(P_{\textit{txt},c})) / \tau})}{\sum_{j=1}^{C} \exp({\cos(I, f_\textit{txt-img}(P_{\textit{txt},j})) / \tau})} &\\
    &- \log \frac{\exp({\cos(f_\textit{img-txt}(I), P_{\textit{txt},c}) / \tau})}{\sum_{j=1}^{C} \exp({\cos(f_\textit{img-txt}(I), P_{\textit{txt},j}) / \tau})} \,,
\end{aligned}
\end{equation}
where $c$ is the index of the ground truth label class for the input image and $\tau$ is the temperature. 

Next, to prevent information loss, we introduce \emph{intra-modal loss}. 
Specifically, we reconstruct the original embedding by mapping the mapped embedding back to its initial modality.
Then, we minimize the $\ell_1$ distance between the reconstructed and original embeddings as follows:
\begin{equation}
\small 
\label{loss: intra}
\begin{aligned}
\ell_{\textit{intra}}(I) = & \|I - f_\textit{txt-img}(f_\textit{img-txt}(I)) \|_1 \\
    & + \sum_{j=1}^{C} \| P_{\textit{txt},j} - f_\textit{img-txt}(f_\textit{txt-img} (P_{\textit{txt},j})) \|_1 \,.
\end{aligned}
\end{equation}
The inter- and intra-modal losses serve to reduce the modality gap between image and text, and to preserve all information during mapping.

\paragraph{Training, Inference, and $\pmb{S_{\textit{GMP}}}$.}
Our overall training objectives are a linear combination of the four losses: $\mathcal{L} = \ell_{\textit{ID}} + \alpha( \ell_{\textit{intra}} + \ell_{\textit{inter}}) + \beta \ell_{\textit{bias}}$,
where $\alpha$ and $\beta$ are trade-off parameters. 
During inference, we use the mean $\mu$ of the image domain bias as the sample $\mathbf{b}$ to generate image domain-biased prompts. 

\textbf{Generalized multi-modal prototypes OOD score $\pmb{S_{\textit{GMP}}}$.}
Building on our empirical and theoretical findings in \cref{Sec: MMA}, with the image-to-text mapping $f_\textit{img-txt}$, we introduce the generalized multi-modal prototypes OOD score $S_{\textit{GMP}}$. 
$S_{\textit{GMP}}$ uses the average of the maximum similarities between the multi-modal input embedding ($I$ and $I' = f_\textit{txt-img}(I)$) and multi-modal prototypes:
\begin{align}
    S_{\textit{GMP}} & (I, I', \{P_{\textit{txt}, c}\}_{c=1}^{C}, \{P_{\textit{img}, c}\}_{c=1}^{C}) =  \nonumber\\
    & \big(S_{\textit{MCM}}(I,\{P_{\textit{txt}, c}\}_{c=1}^{C}) + S_{\textit{MCM}}(I,\{P_{\textit{img}, c}\}_{c=1}^{C})  \label{eq: MMO}\\
    + & S_{\textit{MCM}}(I',\{P_{\textit{txt}, c}\}_{c=1}^{C}) + S_{\textit{MCM}}(I',\{P_{\textit{img}, c}\}_{c=1}^{C})\big)/4  \,.\nonumber
\end{align}
\begin{remark}
    $S_{\textit{GMP}}$ extends $S_{\textit{MCM}}$ by incorporating both uni- and cross-modal similarities. 
    In different scenarios, the importance of intra- and inter-modal similarity is different. 
    However, we \textbf{average the contributions} from the image and text modalities as a \textit{starting point to ensure the generalization ability} of $S_{\textit{GMP}}$ and reduce the gap.
    This approach enhances the separation between ID and OOD samples, improving the OOD detection performance of VLM-based methods, as shown in \cref{fig: ablations: scores}. 
\end{remark}

\begin{table*}[t!]
\resizebox{\textwidth}{!}{%
\begin{tabular}{@{}l|cccccccc|cc@{}}
\toprule
\multirow{3}{*}{Methods} & \multicolumn{8}{c|}{ImageNet-100 $\rightarrow$ OOD Datasets} & \multicolumn{2}{c}{\multirow{2}{*}{Average}} \\
 & \multicolumn{2}{c}{iNaturalist} & \multicolumn{2}{c}{SUN} & \multicolumn{2}{c}{Places} & \multicolumn{2}{c|}{Texture} &  \\ \cmidrule(lr){2-3} \cmidrule(lr){4-5} \cmidrule(lr){6-7} \cmidrule(lr){8-9}
 & FPR95 $\downarrow$ & AUROC $\uparrow$ & FPR95 $\downarrow$ & AUROC $\uparrow$ & FPR95 $\downarrow$ & AUROC $\uparrow$ & FPR95 $\downarrow$ & AUROC $\uparrow$ & FPR95 $\downarrow$ & AUROC $\uparrow$ \\
\midrule
MSP & 23.55 & 95.92 & 37.02 & 92.45 & 40.76 & 91.23 & 24.40 & 94.90 & 31.43 & 93.63 \\
ViM & 20.11 & 96.22 & 38.56 & 93.12 & 44.01 & 87.33 & 33.12 & 93.24 & 33.95 & 92.48 \\
MCM & 18.13 & 96.77 & 36.45 & 94.54 & 34.52 & 94.36 & 41.22 & 92.25 & 32.58 & 94.48 \\
CoOp& 9.30 & 97.95 & 11.64 & 97.61 & 17.45 & 96.53 & 15.94 & 96.90 & 13.58 & \underline{97.25}  \\ %
CoCoOp& 11.76 & 97.84 & 14.28 & 97.13 & 15.16 & 96.73 & 18.27 & 96.54 & 14.86 & 97.06 \\
\midrule
\rowcolor{green!10}\ours & \textbf{2.31} & \textbf{99.28} & \underline{8.26} & \textbf{98.38} & \underline{12.20} & \underline{97.12} & \underline{11.10} & \underline{97.53} & \underline{8.47} & \textbf{98.08} \\
\rowcolor{green!10}\ours{}$^{\dagger}$ & \underline{2.54} & \underline{99.21} & \textbf{8.08} & {\underline{98.28}} & \textbf{12.17} & {\textbf{97.15}} & {\textbf{9.98}} & {\textbf{97.69}} & \textbf{8.19} & \textbf{98.08} \\
\bottomrule
\end{tabular}%
}
\vspace{-0.8em}
\caption{OOD detection results with ID data of ImageNet-100 and four OOD datasets using CLIP (VIT-B/16). Best in \textbf{bold}. 
Second best is \underline{underlined}.
}
\label{tab: main: imagenet100}
\vspace{-1em}
\end{table*}

\section{Experiments}

\subsection{Experimental Details}
\textbf{Datasets and benchmarks.} 
Our experiments primarily use the ImageNet-1k~\cite{dengImageNetLargescaleHierarchical2009} and ImageNet-100~\cite{dengImageNetLargescaleHierarchical2009} dataset as ID data. 
Aligning with standards from prior works~\cite{ming2022delving,wang_clipn_2023}, we evaluate on four diverse OOD datasets: iNaturalist~\cite{van_horn_inaturalist_2018}, SUN~\cite{xiao_sun_2010}, Places~\cite{zhou_places_2018}, and Texture~\cite{cimpoi_describing_2014}. 
OpenOOD benchmark~\cite{zhang_openood_2023} is also used for evaluation on ImageNet-1k.
We test on ImageNet-10~\cite{dengImageNetLargescaleHierarchical2009} and ImageNet-20~\cite{dengImageNetLargescaleHierarchical2009} for hard OOD.  
Details are in the Appendix. 

\noindent
\textbf{Baselines methods.} 
We compare \ours with several baseline methods~\cite{huang_mos_2021,fort_exploring_2021,liu_energy-based_2020,hendrycks_baseline_2017,liang_enhancing_2020,Wang_2022_CVPR,sun_out--distribution_2022,ming2022delving,wang_clipn_2023,jiang2024negative,cao_envisioning_2024,tao_non-parametric_2023,miyai_locoop_2023,liu_category-extensible_2023,nie_out--distribution_2024,bai_id-like_2024,li_learning_2024,yu_self-calibrated_2024,zeng_local-prompt_2025,zhou_learning_2022,zhou2022cocoop}.

\noindent
\textbf{Implementation details.}
For each class, 16 images are used for few-shot tuning and the calculation of ID image prototypes. 
We use the image and text encoders of VIT-B/16 pre-trained by CLIP~\cite{radford_learning_2021} for all the experiments. 
To avoid overfitting, instead of using deep MLPs, we employ linear transformations without bias as the mappings ($f_\textit{img-txt}$ and $f_\textit{txt-img}$). 
The parameters of CLIP are frozen.
We use SGD to optimize other parameters, \eg, estimated image domain bias and learnable context, with a momentum of $0.9$. 
Training epochs, learning rate, batch size, and learnable context length are set to $50, 0.002, 32$, and $16$, respectively. 
We set $\alpha = 0.005$ and $\beta=0.1$. 
Experiments are conducted on a single Nvidia V100 with 32 GB memory. 
All the results for \ours are the average of three trials.

\noindent
\textbf{Metrics.} 
We employ two widely accepted metrics: FPR95, the false positive rate of OOD images when the true positive rate of ID images is at 95\%, and AUROC, the area under the receiver operating characteristic curve.

\subsection{Main Results}

\textbf{ImageNet-1k and ImageNet-100 as the ID dataset}. 
In \cref{tab: main}, we present a comprehensive analysis of ImageNet-1k (ID) across four OOD datasets. 
\ours demonstrates superior performance, achieving the lowest FPR95 and highest AUROC scores across most datasets. 
Notably, \ours achieves an average FPR95 of 20.70 and an average AUROC of 95.38. 
With all the training data for calculating ID prototypes, \ours{}$^{\dagger}$ is further boosted with an average FPR95 of 20.28 and an average AUROC of 95.54. 
In \cref{tab: main: imagenet100}, we observe consistent trends using ImageNet-100 as the ID dataset. 
\ours achieves the best average results, with FPR95 of 8.47 and AUROC of 98.08, showing substantial improvements over other methods. 

\noindent
\textbf{ImageNet-1k on OpenOOD}. 
The results on the OpenOOD benchmark are shown in \cref{tab: openood}. 
Compared to previous methods, \ours achieves the best performance in both near and far OOD scenarios. 
Specifically, \ours achieves an FPR95 of 71.44 and AUROC of 75.95 in the near OOD scenario, and an FPR95 of 20.92 and AUROC of 95.45 in the far OOD scenario.

\noindent
\textbf{ImageNet-10 $\leftrightarrow$ ImageNet-20}. 
To evaluate \ours under challenging conditions where ID and OOD datasets share similar semantics, we follow previous work~\cite{ming2022delving,jiang2024negative} and conduct experiments on ImageNet-10 and ImageNet-20 as shown in \cref{tab: main: imagenet10 20}.
\ours achieves the best performance, with the lowest FPR95 and highest AUROC scores in both scenarios, demonstrating its robustness.

\subsection{Ablation Study}
We present an ablation study using CLIP (ViT-B/16) on ImageNet-1k $\rightarrow$ four OOD datasets. 
The computational complexity analysis (\cref{Appendix: Tab: Parameters}), compatibility with post-hoc methods (\cref{Appendix: tab: post-hoc}), impacts of shot count (\cref{fig: ab: shots}), hyperparameter sensitivity (\cref{fig: ab: alpha,fig: ab: size of base images,fig: ab: length of prompts}), and the use of out-of-distribution prototypes (\cref{Appendix: tab: post-hoc}) are in the Appendix.

\begin{table}[t]
\centering
\resizebox{0.97\columnwidth}{!}{%
\begin{tabular}{l|cc|cc}
\toprule
\multirow{2}{*}{Methods} & \multicolumn{2}{c|}{NearOOD}                     & \multicolumn{2}{c}{FarOOD}                  \\
                         & FPR95 $\downarrow$ & AUROC $\uparrow$ & FPR95 $\downarrow$ & AUROC $\uparrow$\\
\midrule
MCM                      & 85.37                   & 58.97                 & 69.87                   & 77.11                  \\
LoCoOp                   & 87.95                   & 64.97                 & 59.88                   & 83.23         \\
\midrule
\rowcolor{green!10}\ours & \textbf{71.44}          & \textbf{75.95}        & \textbf{20.92}          & \textbf{95.45}      \\
\bottomrule
\end{tabular}%
}
\vspace{-0.9em}
\caption{OOD Detection results with ID data of ImageNet-1k on the OpenOOD benchmark. Best in \textbf{bold}.}
\label{tab: openood}

\centering
\resizebox{\columnwidth}{!}{%
\begin{tabular}{l|cc|cc}
\toprule
\multirow{2}{*}{Methods}    & \multicolumn{2}{c|}{ImageNet-10 $\rightarrow$ ImageNet-20}& \multicolumn{2}{c}{ImageNet-20 $\rightarrow$ ImageNet-10} \\
& FPR95 $\downarrow$   & AUROC $\uparrow$ & FPR95 $\downarrow$   & AUROC $\uparrow$  \\
\midrule
Energy & 10.30 & 97.94 & 16.40 & 97.37\\
CLIPN & 7.80 & 98.07 & 13.67 & 97.47\\
MCM  &         5.00 & 98.71    &     17.40 & 98.87       \\
LoCoOp* & 5.60 & 98.47     & 5.40 & 98.92      \\
\midrule
\rowcolor{green!10}\ours & \textbf{4.10} & \textbf{98.96}  & \textbf{3.60} & \textbf{99.05}\\
\bottomrule
\end{tabular}%
}
\vspace{-0.9em}
\caption{
Comparison on ImageNet-10 $\leftrightarrow$ ImageNet-20 with CLIP-B/16.
Best in \textbf{bold}. 
``*'' represents our reproduction. 
}\label{tab: main: imagenet10 20}
\end{table}

\begin{table}[t!]
\begin{minipage}{.49\columnwidth}
  \centering
  \resizebox{\linewidth}{!}{
    \begin{tabular}{l|cc}
    \toprule
    Methods  & FPR95 $\downarrow$ & AUROC $\uparrow$ \\
    \midrule
    \rowcolor{green!10} \ours & \textbf{20.70} & \textbf{95.38} \\
    \midrule
    - BPG & 28.12&	92.77 \\
    - ITC & 28.74&	93.27 \\
    - Bias in ITC & 24.61 & 94.90 \\
    - $S_{\textit{GMP}}$ & 26.02 & 94.79 \\
    \midrule
    w. MLP & 25.24 & 93.06 \\
    \bottomrule
    \end{tabular}
    }
    \vspace{-0.7em}
    \caption{Effectiveness of different modules and the proposed OOD score $S_{\textit{GMP}}$.}
    \label{tab: ab: modules}
    \vspace{2pt}
\end{minipage}%
\quad
\begin{minipage}{.46\columnwidth}
  \centering
  \resizebox{\linewidth}{!}{
    \begin{tabular}{lc}
    \toprule
    Methods              & Top-1 Acc. (\%) \\
    \midrule
    MCM & 67.0 \\
    CLIPN & 68.5 \\
    ID-like & 68.3 \\
    \midrule
    \rowcolor{green!10} \ours & \textbf{71.8}               \\
    \bottomrule
    \end{tabular}%
    }
    \vspace{-0.7em}
    \caption{Comparison in ID accuracy on ImageNet-1K val set.}
    \label{tab: ab: ID accuracy}
\end{minipage} 
\vspace{-0.2em}
\centering
\resizebox{\linewidth}{!}{%
\begin{tabular}{lccccc}
\toprule
$\left\|\overrightarrow{\Delta}_{\text{gap}}\right\|$ & ImageNet (ID) $\downarrow$ & iNaturalist $\downarrow$ & Places $\downarrow$ & Texture $\downarrow$ & SUN $\downarrow$ \\
\midrule
MCM &  0.8679 & 0.9839 & 0.9101 & 0.9373 & 0.9252 \\
\rowcolor{green!10} \ours & \textbf{0.6135} & \textbf{0.7159}  & \textbf{0.8961} & \textbf{0.9273} & \textbf{0.9002} \\
\bottomrule
\end{tabular}%
}
\vspace{-0.8em}
\caption{Quantifying the modality gap with different methods. We use CLIP (ViT-B/16) as the base model.}
\label{tab: modality gap}
\vspace{-1.3em}
\end{table}

\noindent
\textbf{Effectiveness of BPG, ITC, and $\pmb{S_{\textit{GMP}}}$.} 
In \cref{tab: ab: modules}, we evaluate the effectiveness of BPG and ITC, as well as the proposed score $S_{\textit{GMP}}$, for enhancing OOD detection. 
We see that excluding BPG (-BPG) raises FPR95 to 28.12 and reduces AUROC to 92.77, highlighting BPG’s importance in reducing false positives. 
Similarly, removing ITC (-ITC) significantly degrades performance, increasing FPR95 to 28.74 and lowering AUROC to 93.27. 
We also observe that the Gaussian bias term in ITC is important, as evidenced by the drop in performance (fourth row). 
Furthermore, when the model uses $S_{\textit{MCM}}$ instead of $S_{\textit{GMP}}$, the performance also declines, with FPR95 at 26.02 and AUROC at 94.79. 
Our full model, incorporating ITC, BPG, and $S_{\textit{GMP}}$, achieves the best performance. 
The detailed table is in \cref{tab: ablations: modules full} in the Appendix. 

\noindent
\textbf{Using MLP for \ITC.} 
$f_\textit{img-txt}$ and $f_\textit{txt-img}$ in ITC can sufficiently address the modality gap, we replace them with a two-layer MLP (Linear-ReLU-Linear) and present the experimental results in \cref{tab: ab: modules} (w. MLP).
Results indicate that using MLP negatively impacts performance, which might be because, with limited training data, sophisticated modules such as MLP could overfit.

\noindent
\textbf{\ours improves the ID performance.} 
In \cref{tab: ab: ID accuracy}, we compare the Top-1 accuracy of different methods on the ImageNet-1K validation set. 
Compared to other OOD detection methods, \eg, CLIPN (68.5\%) and ID-like (68.3\%), \ours shows a considerable gain with the highest accuracy, demonstrating its effectiveness in maintaining high ID accuracy while also excelling in OOD detection.

\begin{figure}[t!]
\centering
\centering

    \centering
    \begin{subfigure}[t]{0.48\columnwidth}
        \centering
        \includegraphics[width=\linewidth]{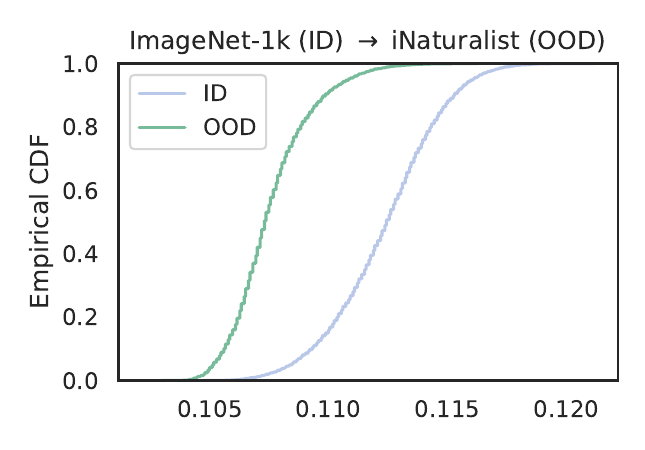}
        \vspace{-1.7em}
        \caption{$S_{\textit{MCM}}, \textit{KS}=0.7270$~\cite{ming2022delving}.}
    \end{subfigure}%
    \begin{subfigure}[t]{0.48\columnwidth}
        \centering
        \includegraphics[width=\linewidth]{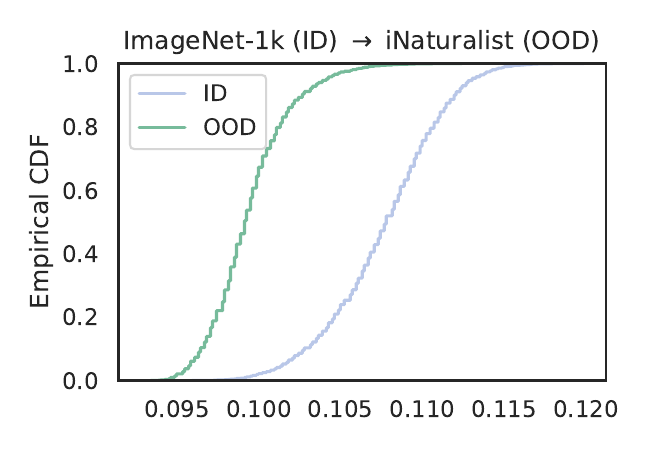}
        \vspace{-1.7em}
        \caption{$S_{\textit{GMP}}, \textit{KS}=0.8555$.}
    \end{subfigure}%
    \vspace{-0.8em}
    \caption{Comparison of (a) $S_{\textit{MCM}}$~\cite{ming2022delving} and (b) our $S_{\textit{GMP}}$ on ImageNet-1k (ID) to iNaturalist (OOD). 
    The scores are multiplied by $100$ for better illustration. 
    \textit{KS} is the Kolmogorov–Smirnov statistic. 
    Higher \textit{KS} values indicate a greater difference between the distributions.
    Best viewed in color. 
    }
    \label{fig: ablations: scores}
    \vspace{-0.2em}
    \centering
    \includegraphics[width=0.9\columnwidth]{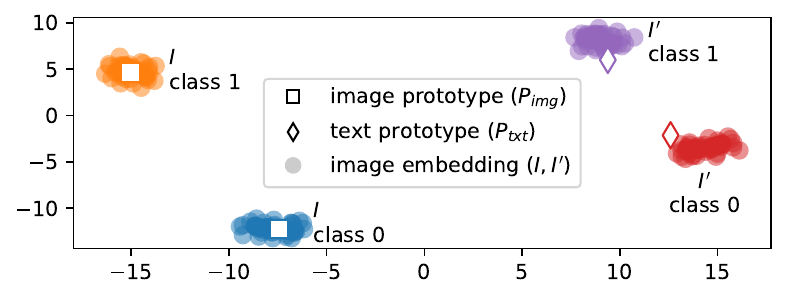}
    \vspace{-1em}
    \caption{The t-SNE visualization of embeddings from two arbitrary classes in ImageNet-1k (Val) using \ours.}
    \label{fig: tsne viz}
    \vspace{-0.5em}
\end{figure}

\noindent
\textbf{Effectiveness of $\pmb{S_{\textit{GMP}}}$.} 
We visualize the empirical cumulative distribution of $S_{\textit{MCM}}$ and our $S_{\textit{GMP}}$ in \cref{fig: ablations: scores} with the ID data of ImageNet-1k and the OOD data of iNaturalist using the same model (\ours). 
$S_{\textit{GMP}}$ clearly improves the gap between ID and OOD data in the empirical CDF figure. 
Furthermore, the Kolmogorov–Smirnov test statistic is enlarged from 0.7270 ($S_{\textit{MCM}}$) to 0.8555 ($S_{\textit{GMP}}$), indicating that using $S_{\textit{GMP}}$ yields more separable scores than $S_{\textit{MCM}}$.

\noindent
\textbf{Reduced modality gap (t-SNE visualization and quantitative assessment of modality gap).}
We choose two arbitrary classes from ImageNet-1k (Val) and use t-SNE to visualize embeddings in \cref{fig: tsne viz}. 
It is clear that ITC reduces the modality gap, as the distances between mapped image embeddings $I'$ and text prototypes $P_\textit{txt}$ are smaller than that between $I$ and $P_\textit{txt}$. 
Similarly, we observe that image embeddings $I$ are located close to image prototypes $P_\textit{img}$. 
We also quantitatively assess the modality gap following Liang et al.~\cite{liang2022mind} as shown in \cref{tab: modality gap,ablation: tab: modality gap}, which shows that \ours achieves smaller modality gaps compared with previous methods.

\section{Conclusion}
We have introduced \ours, a novel VLM-based few-shot tuning OOD detection method.
\ours incorporates two key modules: BPG, which enhances image-text fusion, and ITC, which minimizes the inter- and intra-modal distances. 
We also proposed the generalized multi-modal prototype OOD score, $S_{\textit{GMP}}$, using multi-modal prototypes and embeddings to improve robustness and reduce false positives. 
Experiments show that \ours consistently outperforms existing VLM-based OOD detection methods. 

\noindent
\textbf{Limitations}. 
In some scenarios, the ID data may not be available.
Exploring methods such as image generation or retrieval from a web-scale dataset could help alleviate the dependency on the ID data. 
Moreover, exploring more advanced distribution models rather than the Gaussian distribution we use may further improve the performance.

\section*{Acknowledgement}

This work was supported by the Natural Sciences and Engineering Research Council of Canada (NSERC)-CSE Research Community project entitled ``An End-to-End Approach to Safe and Secure AI Systems''. 

Researchers funded through the NSERC-CSE Research Communities Grants do not represent the Communications Security Establishment Canada or the Government of Canada. 
Any research, opinions, or positions they produce as part of this initiative do not represent the official views of the Government of Canada.

{
    \small
    \bibliographystyle{ieeenat_fullname}
    \bibliography{main,references}
}

\clearpage
\setcounter{page}{1}
\maketitlesupplementary

\begin{figure*}[h!]
    \centering
    \subfloat[][ImageNet $\rightarrow$ iNaturalist (OOD)]{\includegraphics[width=0.48\linewidth]{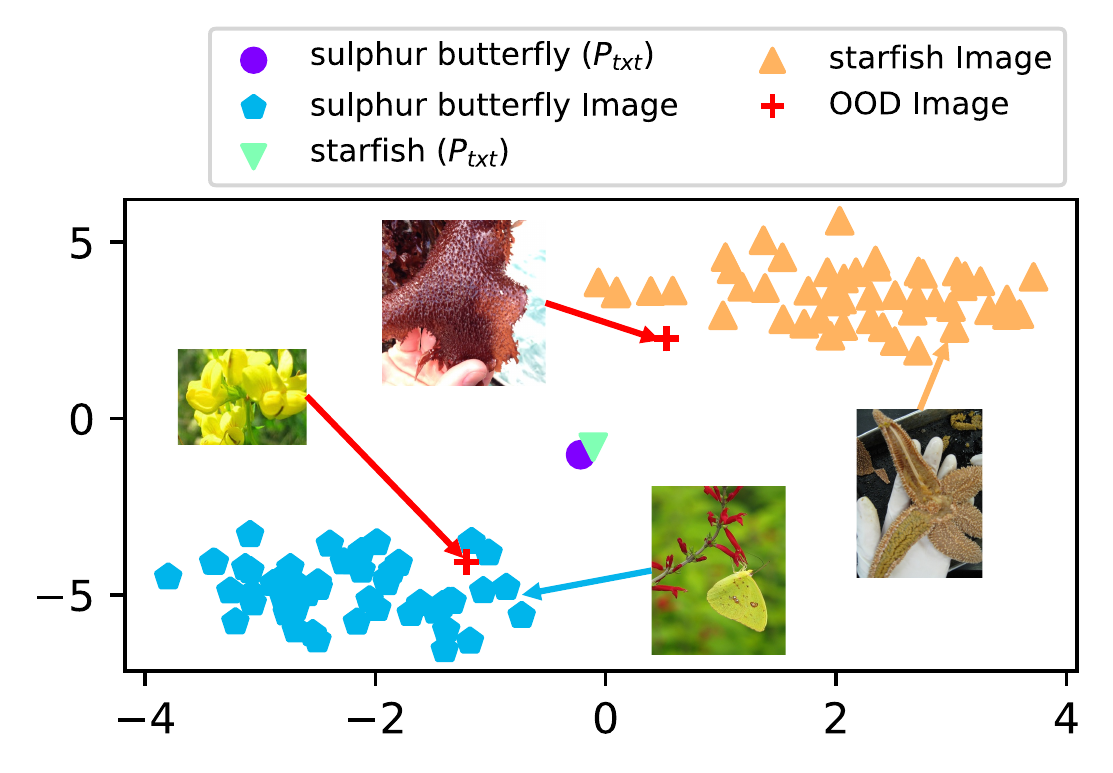}}
    \subfloat[][ImageNet $\rightarrow$ Texture (OOD)]{\includegraphics[width=0.48\linewidth]{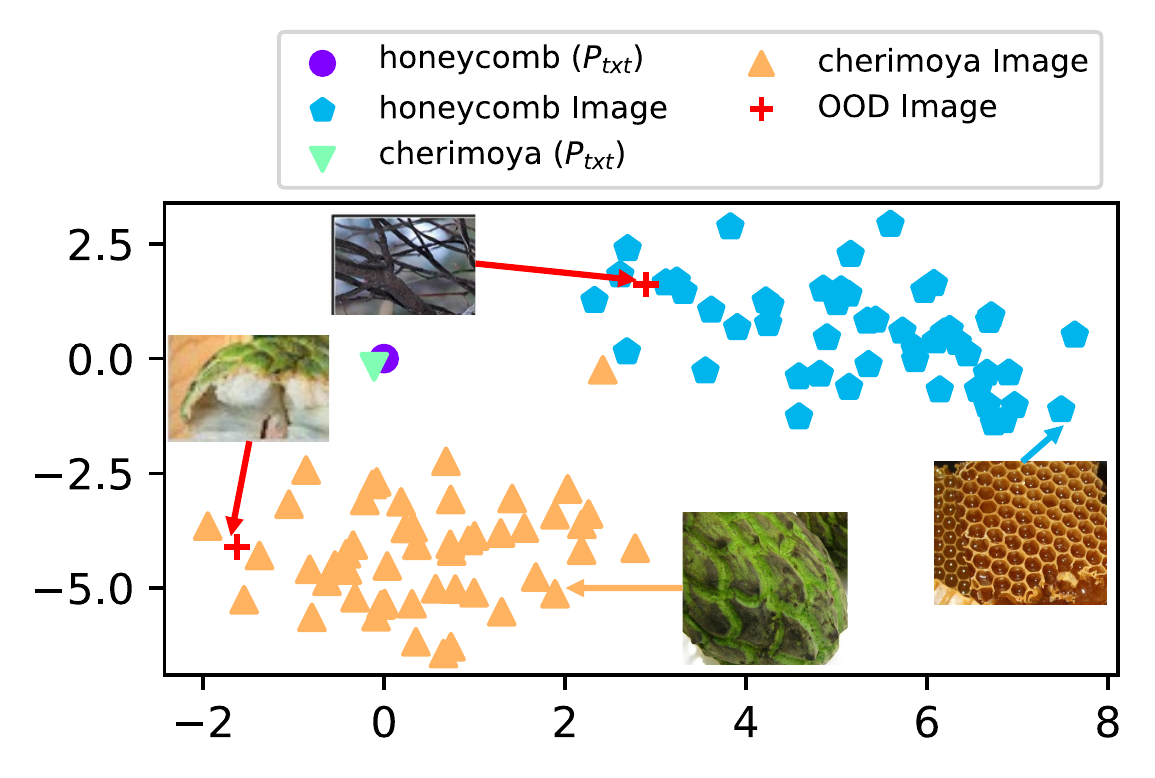}}

    \subfloat[][ImageNet $\rightarrow$ Places (OOD)]{\includegraphics[width=0.48\linewidth]{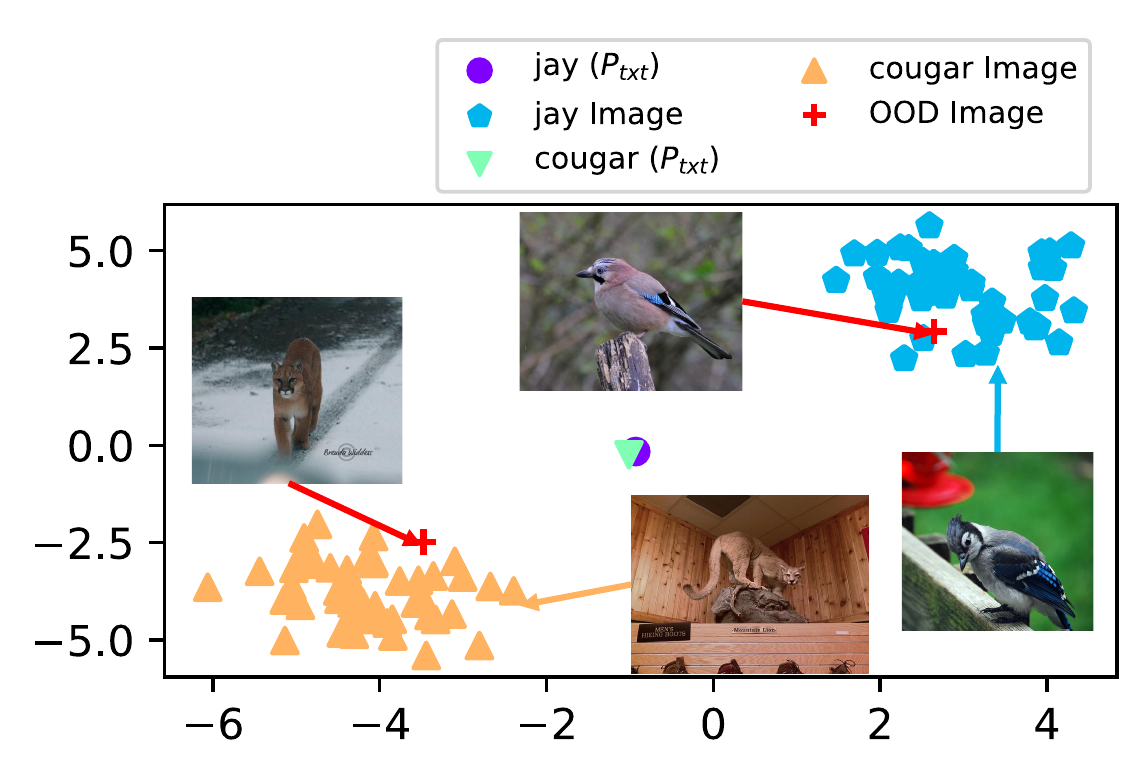}}
    \subfloat[][ImageNet $\rightarrow$ SUN (OOD)]{\includegraphics[width=0.48\linewidth]{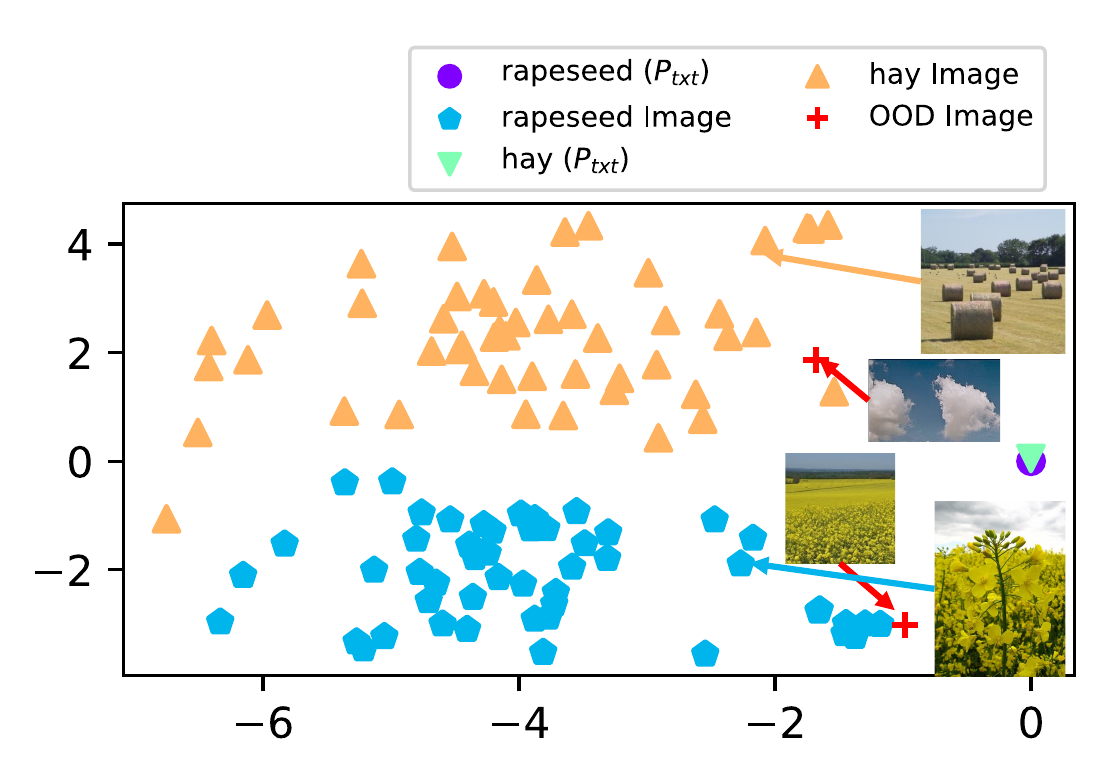}}

    \caption{
    Examples where OOD images from four OOD datasets are located closer to ID text prototypes compared to ID images (ImageNet-1k) generated with CLIP (ViT-B/16).}
    \label{ablations: fig: ood example}
\end{figure*}

In this Appendix, we present the details and complete results corresponding to \cref{tab: MMA} in \cref{Appendix: Sec: full MMA}, full theoretical motivation in \cref{appendix: sec: theoretical}, additional ablation studies in \cref{Appendix: Sec: More ablations}, benchmark dataset specifications in \cref{Appendix: Sec: datasets}, and the proof of \cref{the: image anchor helps} in \cref{Appendix: Sec: Proof}.

\section{OOD Examples}

In \cref{fig: ood example,ablations: fig: ood example}, we visualize the embedding space using t-SNE projection across four different OOD datasets (iNaturalist, Texture, Places, and SUN) generated with CLIP (ViT-B/16). 
A pattern emerges: OOD images (red crosses) are consistently positioned closer to ID text prototypes than their corresponding ID images across all four scenarios. 
For instance, in \cref{ablations: fig: ood example}(a), OOD images (red cross) are positioned closer to the ``sulphur butterfly'' text prototype (purple circle) than the corresponding ID images (blue pentagon). 
This counterintuitive positioning is systematically observed across diverse semantic categories—from biological entities (sulphur butterfly, starfish) to textures (honeycomb, cherimoya) and environmental scenes (jay/cougar, rapeseed/hay).

This phenomenon reveals a fundamental limitation in current embedding spaces: \textit{positional proximity does not guarantee semantic similarity}. 
The t-SNE visualization demonstrates that geometric distance in the learned representation space can be misleading, as semantically unrelated OOD samples may appear closer to class prototypes than genuine in-distribution examples. 
This misalignment between spatial proximity and semantic similarity undermines the reliability of distance-based OOD detection methods. 
It motivates our design of multimodal prototypes that better capture the true semantic boundaries of in-distribution data.

\section{RQ1}

\subsection{More Empirical Evidence for RQ1}
\label{Appendix: Sec: full MMA}

\begin{table}[h!]
\centering
\resizebox{\columnwidth}{!}{%
\begin{tabular}{l|cc}
\toprule
Methods    & FPR95 (\%) $\downarrow$   & AUROC (\%) $\uparrow$  \\
\midrule 
\rowcolor[gray]{.9} \multicolumn{3}{c}{ImageNet-100 $\rightarrow$ 4 datasets (Average)} \\
Text prototypes        &    32.58  &94.48  \\
Image prototypes & 31.09 & 94.74 \\
Text + Image prototypes        & 24.18  & 95.79      \\
\midrule
\rowcolor[gray]{.9} \multicolumn{3}{c}{ImageNet-100 $\rightarrow$ iNaturalist} \\
Text prototypes        &    18.13 & 96.77          \\
Image prototypes & 26.06 & 93.09 \\
Text + Image prototypes        &    \textbf{14.76} & \textbf{97.36}         \\
\rowcolor[gray]{.9} \multicolumn{3}{c}{ImageNet-100 $\rightarrow$ SUN} \\
Text prototypes        &    36.45 & \textbf{94.54}        \\
Image prototypes & 39.25 & 93.17 \\
Text + Image prototypes        &    \textbf{30.28} & 92.51       \\
\rowcolor[gray]{.9} \multicolumn{3}{c}{ImageNet-100 $\rightarrow$ Places} \\
Text prototypes        &    34.52 & \textbf{94.36}        \\
Image prototypes & 40.59 & 96.43 \\
Text + Image prototypes        &    \textbf{34.04} &    93.92      \\
\rowcolor[gray]{.9} \multicolumn{3}{c}{ImageNet-100 $\rightarrow$ Texture} \\
Text prototypes        &    41.22 & 92.25        \\
Image prototypes & 18.46 & 96.27 \\
Text + Image prototypes        &    \textbf{17.66} & \textbf{96.66}        \\
\bottomrule
\end{tabular}%
}
\caption{
Comparison of OOD detection results on four OOD tasks. 
We use CLIP-B/16 for $S_{\textit{MCM}}$ and $S_{\textit{MMP}}$ (ours, \cref{eq: MMA}). 
FPR95 represents the false positive rate of OOD images when the true positive rate of ID images is at 95\% while AUROC is the area under the
receiver operating characteristic curve.
Best in \textbf{bold}. 
}\label{tab: MMA full}
\end{table}

To empirically justify the effectiveness of multi-modal (image and text) prototypes, we present experiments using $S_{\textit{MCM}}$ and $S_{\textit{MMP}}$ on four OOD datasets (\ie, iNaturalist, SUN, Places, and Texture) as shown in \cref{tab: MMA full}. 
We use CLIP-B/16 as the base model for $S_{\textit{MCM}}$ and $S_{\textit{MMP}}$. 
All the training images from ImageNet-100 are used for generating image prototypes. 
We also include $S_{\textit{MCM}}$ with image prototypes as a comparison baseline.

Our method, $S_{\textit{MMP}}$, consistently demonstrates superior performance in terms of FPR95 across all four tasks, achieving the lowest values, with notable improvements on iNaturalist (14.76 vs. 18.13) and Texture (17.66 vs. 41.22), highlighting its robustness in reducing false positives. 
For AUROC, $S_{\textit{MMP}}$ achieves the highest score on iNaturalist (97.36) and Texture (96.66). 
However, in the SUN and Places datasets, while $S_{\textit{MMP}}$ maintains competitive performance, it slightly underperforms $S_{\textit{MCM}}$ in AUROC (92.51 vs. 94.54 on SUN and 93.92 vs. 94.36 on Places). 

Similarly, we observe that $S_{\textit{MMP}}$ outperforms $S_{\textit{MCM}}$ with image prototypes on most of the metrics, while $S_{\textit{MCM}}$ with image prototypes has slightly better performance compared to $S_{\textit{MCM}}$ (with text prototypes). 
This shows that the modality gap negatively impacts the performance.

Overall, the results underscore the effectiveness of $S_{\textit{MMP}}$ and multi-modal prototypes in reducing false positives and the performance of VLM-based OOD detection.

\subsection{Theoretical Evidence for RQ1}\label{appendix: sec: theoretical}

To further understand why image prototypes improve performance, we present a theoretical analysis in \cref{the: image anchor helps}. 

\begin{theorem}[Multi-modal Prototypes Increase Score Separation between ID and OOD Data]\label{the: image anchor helps} 
    Assuming that the %
    OOD data is not drawn from any ID distribution, we have,
    \begin{equation}
    \small
        \begin{aligned}
            & \mathbb{E}\left[ S_{\textit{MMP}}(I_{\textit{ID}}) - S_{\textit{MMP}}(I_{\textit{OOD}}) \right]             \geq \mathbb{E}\left[ S_{\textit{MCM}}(I_{\textit{ID}}) - S_{\textit{MCM}}(I_{\textit{OOD}}) \right]\,,
        \end{aligned}
    \end{equation}
    where $I_{\textit{ID}}$ and $I_{\textit{OOD}}$ are the image embeddings of ID and OOD samples. We omit multi-modal prototypes for clarity. %
\end{theorem}

This greater score separation enhances the separation between ID and OOD data, improving detection performance, under general assumptions used by previous works~\cite{fu_clipscope_2025,liang2022mind}.
And it further improves performance. 
The proof is provided in \cref{Appendix: Sec: Proof}.

\section{Experiments}

\begin{table}[t!]
    \centering
    \begin{tabular}{l|cc}
    \toprule
    Methods & FPR95 (\%) $\downarrow$ & AUROC (\%) $\uparrow$ \\
    \midrule
    LoCoOp                   & 85.60                   & 60.73                 \\
    \rowcolor{green!10}\ours & \textbf{64.97}          & \textbf{75.32}       \\
    \bottomrule
    \end{tabular}%
    \caption{OOD Detection results with ID data of ImageNet-1k on the OpenOOD benchmark (CSID). Best in \textbf{Bold}.}
    \label{tab: openood csid}
\end{table}

\begin{table*}[t!]
\centering
\resizebox{\textwidth}{!}{%
\begin{tabular}{@{}ccc|cccccccc|cc@{}}
\toprule
\multirow{3}{*}{ITC} & \multirow{3}{*}{BPG} & \multirow{3}{*}{$S_{\textit{GMP}}$} & \multicolumn{8}{c}{ImageNet-1k $\rightarrow$ OOD Datasets} & \multicolumn{2}{|c}{\multirow{2}{*}{Average}} \\
 & & & \multicolumn{2}{c}{iNaturalist} & \multicolumn{2}{c}{SUN} & \multicolumn{2}{c}{Places} & \multicolumn{2}{c|}{Texture} &  \\ \cmidrule(lr){4-5} \cmidrule(lr){6-7} \cmidrule(lr){8-9} \cmidrule(lr){10-11}
 &  & & FPR95 $\downarrow$ & AUROC $\uparrow$ & FPR95 $\downarrow$ & AUROC $\uparrow$ & FPR95 $\downarrow$ & AUROC $\uparrow$ & FPR95 $\downarrow$ & AUROC $\uparrow$ & FPR95 $\downarrow$ & AUROC $\uparrow$ \\
\midrule
&&& 30.91 & 94.61 & 37.59 & 92.57 & 44.69 & 89.77 & 57.77 & 86.11 & 42.74 & 90.77 \\
\checkmark& \checkmark && 8.56 & 98.48 & 22.36 & 95.38 & 33.60 & 92.55 & 39.57 & \textbf{92.75} & 26.02 & 94.79 \\
\checkmark& & \checkmark & 20.86 & 95.06 & 22.04 & 94.92 & 30.22 & 91.64 & 39.38 & 89.46 & 28.12 & 92.77 \\
&\checkmark&\checkmark& 16.10 & 96.50 & 23.81 & 94.53 & 32.63 & 91.94 & 42.41 & 90.11 & 28.74 & 93.27 \\
\checkmark & \checkmark (w/o Bias)  & \checkmark  & 21.55 & 97.79 & 16.37 & \textbf{98.00} & 26.25 & 92.24 & 34.25 & 91.56 & 24.61 & 94.90 \\
\midrule
\rowcolor{green!10}\checkmark & \checkmark  & \checkmark  & \textbf{5.91} & \textbf{98.73} & \textbf{15.85} & {96.08} & {\textbf{26.18}} & \textbf{93.19} & \textbf{34.88} & 91.93 & \textbf{20.70} & \textbf{95.38} \\
\midrule
\multicolumn{3}{c|}{w. ResNet-50} & 21.93 & 96.22 & 37.73 & 91.00 & 43.27 & 87.41 & 28.14 & 93.51 & 32.77 & 92.04 \\
\multicolumn{3}{c|}{w. ResNet-101} & 6.13 & 98.58 & 20.61 & 95.03 & 33.16 & 90.94 & 41.31 & 90.70 & 25.30 & 93.81 \\
\bottomrule
\end{tabular}%
}
\caption{
Effectiveness of different modules, the proposed OOD score $S_{\textit{GMP}}$, and different backbones. 
When $S_{\textit{GMP}}$ is not combined with BPG, it degenerates to $S_{\textit{MCM}}$.
}
\label{tab: ablations: modules full}
\end{table*}

\begin{table*}[t!]
\centering
\resizebox{\textwidth}{!}{%
\begin{tabular}{cc|cccccccc|cc}
\toprule
\multirow{3}{*}{ReAct} & \multirow{3}{*}{$S_{\textit{EOE}}$} & \multicolumn{8}{c|}{ImageNet-1k $\rightarrow$ OOD Datasets}                                                                                                   & \multicolumn{2}{c}{\multirow{2}{*}{Average}} \\
                       &                           & \multicolumn{2}{c}{iNaturalist}       & \multicolumn{2}{c}{SUN}               & \multicolumn{2}{c}{Places}            & \multicolumn{2}{c|}{Texture}          & \multicolumn{2}{c}{}                         \\
                       &                           & FPR95 $\downarrow$ & AUROC $\uparrow$ & FPR95 $\downarrow$ & AUROC $\uparrow$ & FPR95 $\downarrow$ & AUROC $\uparrow$ & FPR95 $\downarrow$ & AUROC $\uparrow$ & FPR95 $\downarrow$     & AUROC $\uparrow$    \\
\midrule
                       &                      & {5.91} & \textbf{98.73} & {15.85} & {96.08} & {{26.18}} & 93.19 & 34.88 & 91.93 & {20.70} & {95.38}       \\
                \checkmark &                      & 8.03 & 97.93 & 17.39 & 95.51 & 24.66 & 93.32 & \textbf{33.21} & \textbf{92.13} & 20.82 & 94.72   \\
                       &   \checkmark                    & \textbf{4.85} & 98.71 & 14.90 & 96.13 & 23.82 & 93.68 & 34.96 & 91.22 & 19.63 & 94.93             \\
             \checkmark           &    \checkmark                   & 7.27 & 98.32 & \textbf{11.79} & \textbf{97.23} & \textbf{20.28} & \textbf{94.36} & 38.87 & 90.97 & \textbf{19.55} & \textbf{95.22} \\
\bottomrule
\end{tabular}%
}
\caption{Performance of \ours combined with post-hoc methods, \ie, ReAct~\cite{sun_react_2021} and $S_{\textit{EOE}}$~\cite{cao_envisioning_2024}.}
\label{Appendix: tab: post-hoc}
\end{table*}

\begin{table}[t!]
\centering
\resizebox{\columnwidth}{!}{%
\begin{tabular}{l|cc|cc}
\toprule
\multirow{2}{*}{Methods} & \multicolumn{2}{c|}{ResNet-50}         & \multicolumn{2}{c}{ResNet-101}        \\
                         & FPR95 $\downarrow$ & AUROC $\uparrow$ & FPR95 $\downarrow$ & AUROC $\uparrow$ \\
\midrule
NegPrompt                & 57.19              & 85.70            & 50.35              & 88.04            \\
LoCoOp                   & 34.45              & 91.95            & -                  & -                \\
\midrule
\rowcolor{green!10} \ours                    & \textbf{30.55}              & \textbf{92.90}            & \textbf{25.30}     & \textbf{93.81}   \\
\bottomrule
\end{tabular}
}
\caption{Performance comparison using different base models.}
\label{tab: different base}
\end{table}

\begin{table*}
\centering
\begin{tabular}{lcccc}
\toprule
Methods   & Parameters                      & Trainable Parameters & Image Prototypes & Inference \\
\midrule
MCM    & 124,323,841                     & 0                    & -                             &    $O(1)$  \\
\ours & 124,891,683 ($\uparrow$567,842) & 567,842              & $O(N_{base})$                           &   $O(1)$  \\
\bottomrule
\end{tabular}%
\caption{Comparison of (trainable) parameters and inference efficiency between MCM and \ours with the base model as CLIP-B/16.}
\label{Appendix: Tab: Parameters}
\end{table*}

\subsection{Results on OpenOOD (CSID)}
We evaluate the performance of \ours on the OpenOOD benchmark (CSID)~\cite{huang_mos_2021} using the ImageNet-1k dataset as the ID data in \cref{tab: openood csid}.
The results show that \ours outperforms LoCoOp, achieving the lowest FPR95 and highest AUROC, demonstrating its effectiveness in OOD detection across different benchmarks.

\subsection{Ablations}
\label{Appendix: Sec: More ablations}

To understand the effectiveness of \ours, we present the results of several ablation studies using CLIP (ViT-B/16) on ImageNet-1k $\rightarrow$ four OOD datasets (\ie, iNaturalist, SUN, Places, and Texture). Our experiments examine the effectiveness of the proposed components (\ie, ITC, BPG, and $S_{\textit{GMP}}$), the impact of combining \ours with post-hoc methods, training data efficiency (number of shots), the statistical evaluation of $S_{\textit{GMP}}$, and sensitivity to hyperparameters.

\noindent
\textbf{Evaluating the effectiveness of the proposed modules (BPG and ITC) and $\pmb{S_{\textit{GMP}}}$ (full results).}
To evaluate the effectiveness of the proposed modules and $S_{\textit{GMP}}$, we present detailed results of a combinatorial ablation study in \cref{tab: ablations: modules full}. 
Performance is evaluated across four datasets (\ie, iNaturalist, SUN, Places, and Texture).

\noindent
\textbf{Combining \ours with post-hoc methods.} 
To understand how \ours combines with existing representative post-hoc methods, \ie, ReAct~\cite{sun_react_2021} and $S_{\textit{EOE}}$~\cite{cao_envisioning_2024}, we present a systematic study in \cref{Appendix: tab: post-hoc}. 
We directly apply ReAct and $S_{\textit{EOE}}$ on our best model as post-hoc methods, which means we do not use them during training. 
For ReAct, we set the threshold at 0.95, while for $S_{\textit{EOE}}$, we follow the standard parameters. 
We see that using ReAct alone makes the performance worse with an average FPR95 of 20.82 and AUROC of 94.72. 
On the other hand, $S_{\textit{EOE}}$ boosts average FPR95 from 20.70 to 19.63. 
When combining ReAct and $S_{\textit{EOE}}$, the FPR95 is boosted to 19.55 on average. 
However, using ReAct and/or $S_{\textit{EOE}}$ decreases the average AUROC compared to vanilla \ours, showing that these post-hoc methods might only help reduce false positives. 
The experiments show that \ours is compatible with these post-hoc methods without performance drops.

\noindent
\textbf{Effectiveness of \ours with different backbones.} 
The results in \cref{tab: different base} show that \ours consistently outperforms other methods across different backbones, including ResNet-50~\cite{heDeepResidualLearning2016} and ResNet-101~\cite{heDeepResidualLearning2016}.
Specifically, \ours achieves 30.55 for FPR95 and 92.90 for AUROC with ResNet-50, and 25.30 for FPR95 and 93.81 for AUROC with ResNet-101, demonstrating its robustness and effectiveness across various model architectures.

\noindent
\textbf{Computational cost of the training and inference.}
Table~\ref{Appendix: Tab: Parameters} compares parameter efficiency and inference scalability between the baseline MCM method and our proposed \ours, using CLIP-B/16 as the base model. 
While both methods exhibit comparable total parameter sizes, \ours introduces 567,842 trainable parameters to achieve its enhanced functionality, representing a negligible increase of 0.46\% in total parameters. 
\ours obtains image prototypes with a complexity of $O(N_{base})$, while maintaining inference complexity at $O(1)$ given one test input. 
Also, the calculation of image prototypes can be conducted before inference, and image prototypes can be shared among all the test samples, which avoids introducing overhead during inference compared with MCM. 
It demonstrates the scalability of \ours without computational overhead during inference, effectively addressing challenges of learnable flexibility and inference efficiency.

\begin{figure}
    \centering
    \includegraphics[width=0.9\columnwidth]{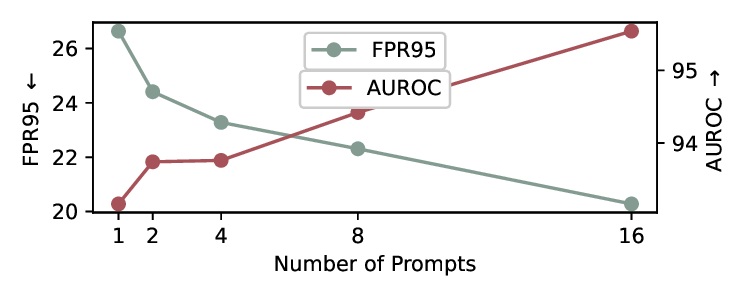}
    \caption{Performance comparison on the length of prompts.}
    \label{fig: ab: length of prompts}
\end{figure}

\begin{figure}
    \centering
    \includegraphics[width=\columnwidth]{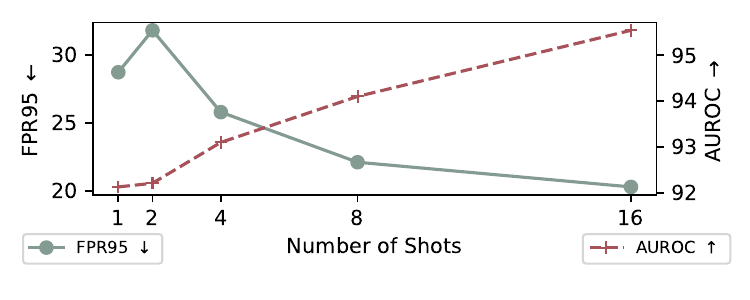}
    \caption{The impact of different sizes of fine-tuning data.}
    \label{fig: ab: shots}
\end{figure}

\begin{figure}[t!]
\centering
\includegraphics[width=\columnwidth]{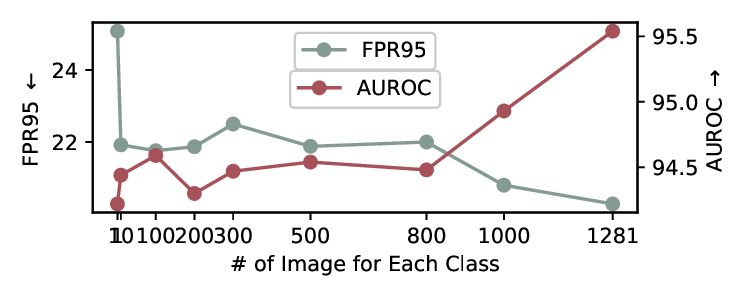}
\caption{Performance comparison on the number of based images used for obtaining image prototypes. ``1281'' corresponds to using all the training images as the base images.}
\label{fig: ab: size of base images}
\end{figure}

\begin{figure}
    \centering
    \includegraphics[width=\columnwidth]{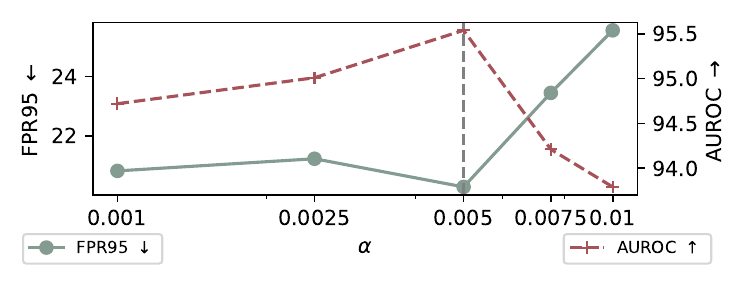}
    \includegraphics[width=\columnwidth]{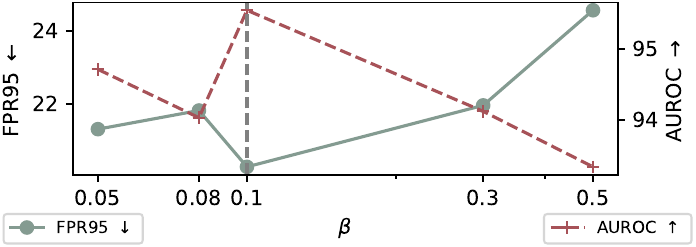}
    \caption{Hyperparameter sensitivity ($\alpha$ and $\beta$). We report the average FPR95 and AUROC.}
    \label{fig: ab: alpha}
\end{figure}

\begin{figure*}
    \centering
    \begin{subfigure}[t]{0.48\columnwidth}
        \centering
        \includegraphics[width=\linewidth]{pics//score_comparison/MCM_imagenet_iNaturalist_ecdf.pdf}
        \caption{$S_{\textit{MCM}}, \textit{KS}=0.7270$~\cite{ming2022delving}.}
    \end{subfigure}
    \begin{subfigure}[t]{0.48\columnwidth}
        \centering
        \includegraphics[width=\linewidth]{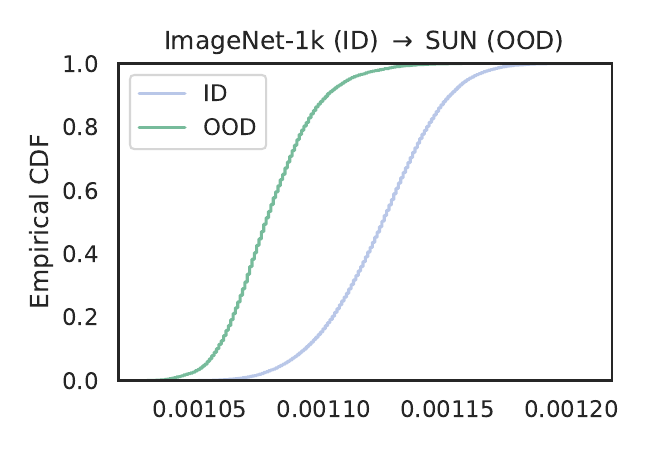}
        \caption{$S_{\textit{MCM}}, \textit{KS}=0.7182$~\cite{ming2022delving}.}
    \end{subfigure}%
    \begin{subfigure}[t]{0.48\columnwidth}
        \centering
        \includegraphics[width=\linewidth]{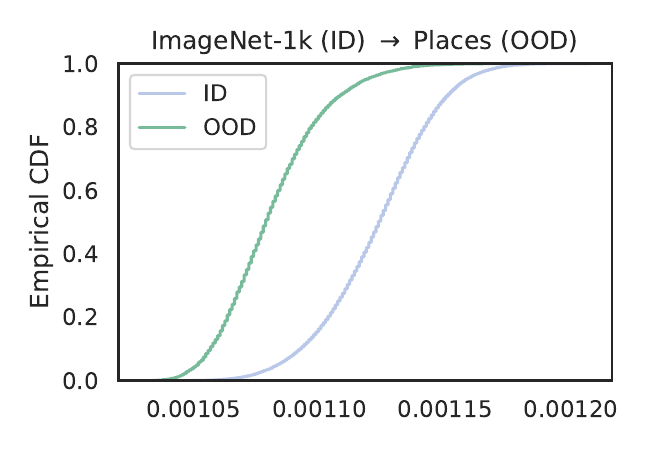}
        \caption{$S_{\textit{MCM}}, \textit{KS}=0.6650$~\cite{ming2022delving}.}
    \end{subfigure}%
    \begin{subfigure}[t]{0.48\columnwidth}
        \centering
        \includegraphics[width=\linewidth]{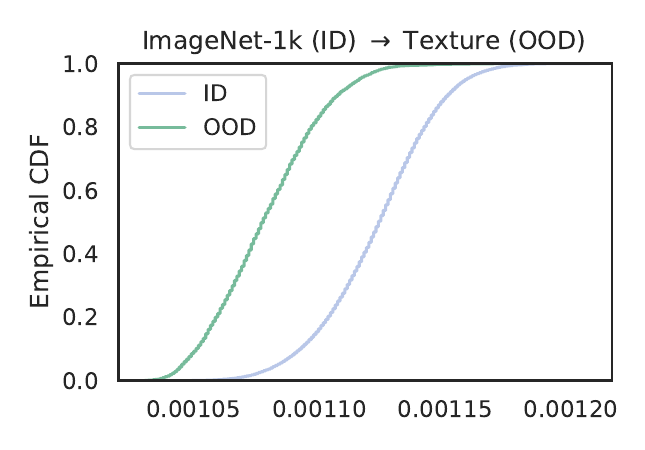}
        \caption{$S_{\textit{MCM}}, \textit{KS}=0.6760$~\cite{ming2022delving}.}
    \end{subfigure}%
    
    \begin{subfigure}[t]{0.48\columnwidth}
        \centering
        \includegraphics[width=\linewidth]{pics/score_comparison/MMO_imagenet_iNaturalist_ecdf.pdf}
        \caption{$S_{\textit{GMP}}, \textit{KS}=0.8555$.}
    \end{subfigure}
    \begin{subfigure}[t]{0.48\columnwidth}
        \centering
        \includegraphics[width=\linewidth]{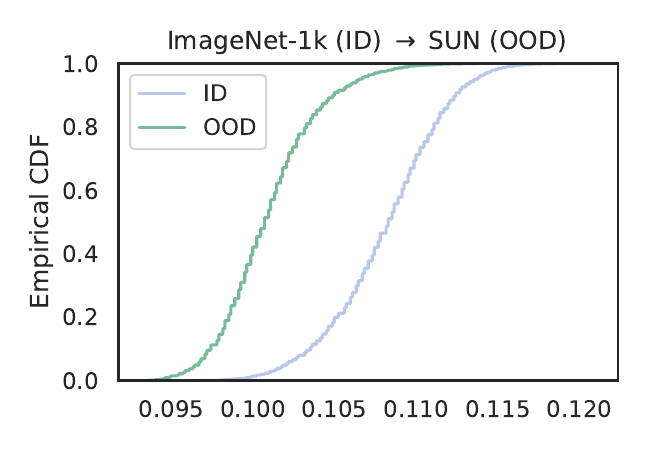}
        \caption{$S_{\textit{GMP}}, \textit{KS}=0.7732$.}
    \end{subfigure}
    \begin{subfigure}[t]{0.48\columnwidth}
        \centering
        \includegraphics[width=\linewidth]{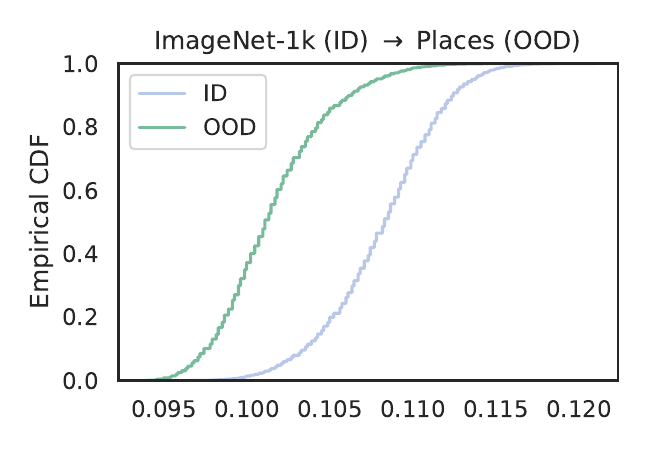}
        \caption{$S_{\textit{GMP}}, \textit{KS}=0.7281$.}
    \end{subfigure}
    \begin{subfigure}[t]{0.48\columnwidth}
        \centering
        \includegraphics[width=\linewidth]{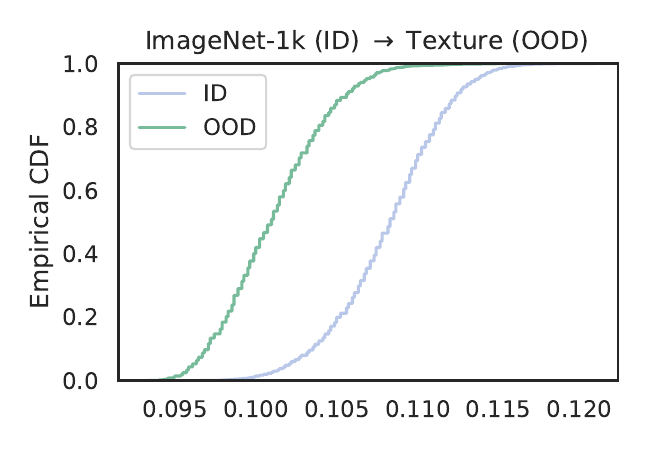}
        \caption{$S_{\textit{GMP}}, \textit{KS}=0.6848$.}
    \end{subfigure}
    
    \caption{The comparison between $S_{\textit{MCM}}$~\cite{ming2022delving} and our proposed $S_{\textit{GMP}}$ score (\cref{eq: MMO}) on ImageNet-1k (ID) to four OOD datasets, \ie, iNaturalist, SUN, Places, and Texture.
    Best viewed in color.
    }
    \label{fig: ablations: scores full}
\end{figure*}

\begin{table*}[t]
\centering
\begin{tabular}{l|cc|ccccc}
\toprule
$\left\|\overrightarrow{\Delta}_{\text{gap}}\right\|$ & Modality 1 & Modality 2& ImageNet (ID) $\downarrow$ & iNaturalist $\downarrow$ & Places $\downarrow$ & Texture $\downarrow$ & SUN $\downarrow$ \\
\midrule
MCM & ${P_{\text{txt}}}$ & $I$ & 0.8679 & 0.9839 & 0.9101 & 0.9373 & 0.9252 \\
\midrule
\multirow{3}{*}{\ours} & ${P_{\text{txt}}}$ & $I$ & 0.6135 & 0.7159  & 0.8961 & 0.9273 & 0.9002 \\
 & ${P_{\text{txt}}}$ & $f_{\text{img-txt}}(I)$ & 0.1938 & \textbf{0.2238}  & \textbf{0.2376} & \textbf{0.2559} & \textbf{0.2454} \\
 & ${P_{\text{img}}}$ & $I$ & \textbf{0.1348} & 0.4531  & 0.3474 & 0.3333 & 0.3447 \\
\bottomrule
\end{tabular}%
\caption{Quantifying the modality gap with different methods. We use CLIP (ViT-B/16) as the base model.}
\label{ablation: tab: modality gap}
\end{table*}

\noindent
\textbf{Impact of the length of prompts.}
In the main experiments, we follow previous works~\cite{miyai_locoop_2023,zhou2022cocoop,zhou_learning_2022} and set the length of prompts to $L = 16$. 
Here, we would like to validate that $16$ is the best choice for \ours. 
We conducted experiments with \ours$^{\dagger}$ and results shown in \cref{fig: ab: length of prompts} indicate that among the choices from $1$ to $16$, $16$ obtains the lowest FPR95 and highest AUROC. 
Increasing the length of prompts improves the OOD detection performance.

\noindent
\textbf{How much data do we need for few-shot tuning?}
We evaluate the effect of the number of shots for few-shot tuning on the model's performance in \cref{fig: ab: shots} with \ours$^{\dagger}$. 
As the number of shots increases from \(1\) to \(16\), we observe a consistent improvement in both metrics. Specifically, FPR95 decreases from 28.73 at \(1\)-shot to 20.28 at \(16\)-shot, indicating that more shots lead to better separability between ID and OOD data. 
Similarly, AUROC improves from 92.13 at \(1\)-shot to a peak of 95.54 at \(16\)-shot, demonstrating enhanced overall detection performance. 
These results highlight the robustness of \ours in leveraging additional labeled examples, with diminishing returns on FPR95 as data size increases. 
While higher shot numbers provide the best performance, the model performs reasonably well even with a limited number of shots, showcasing its flexibility in low-data scenarios.

\noindent
\textbf{Impact of base image size for constructing ID image prototypes.}
In $S_{\textit{GMP}}$, we use a set of base images to construct image prototypes $\{P_{i,c}\}_{c=1}^{C}$. 
To understand how the size of base images affects performance, we conduct experiments varying its size by randomly selecting a subset of the training images.
The results in \cref{fig: ab: size of base images} show that by increasing the number of images for each class, the performance of \ours is improved, as image prototypes are closer to the real distribution. 
FPR95 and AUROC reach the best when using all the images. 
However, as shown in \cref{tab: main}, the performance is already competitive when using the same set of few-shot data for fine-tuning, indicating that the model is robust to the size of base images.

\noindent
\textbf{Hyperparameter ($\alpha$ and $\beta$) sensitivity}.  
We analyze the effect of the trade-off hyperparameters $\alpha$ and $\beta$ (\cref{sec: our method}) on the model's performance, as shown in \cref{fig: ab: alpha} with \ours$^{\dagger}$. 
For $\alpha$, we observe a trade-off between FPR95 and AUROC as its value increases. 
Specifically, as $\alpha$ grows from 0.001 to 0.01, FPR95 decreases initially from 20.82 to 20.28 at $\alpha = 0.005$, indicating improved separability, but then rises to 25.56 at $\alpha = 0.01$. 
Meanwhile, AUROC peaks at 95.54 for $\alpha = 0.005$ before declining, indicating that excessively large values of $\alpha$ may overemphasize certain features, leading to diminished overall performance.  
Similarly, $\beta$ demonstrates a comparable trend. The FPR95 achieves its lowest value of 20.28 at $\beta = 0.1$, while AUROC peaks at 95.54. 
The results of both hyperparameters suggest that \ours is not sensitive to the parameters and that moderate values ($\alpha = 0.005, \beta = 0.1$) balance the trade-off between false positive rate and overall discriminative ability.

\noindent
\textbf{Full statistical evaluation of $\pmb{S_{\textit{GMP}}}$.} 
We visualize the empirical cumulative distribution (empirical CDF) and the Kolmogorov–Smirnov (KS) test statistics for \( S_{\textit{MCM}} \) and our proposed \( S_{\textit{GMP}} \) in \cref{fig: ablations: scores full}.  
The evaluation uses our best model trained on the ImageNet-1k (ID) dataset, with iNaturalist, SUN, Places, and Texture serving as OOD datasets.  
The empirical CDF clearly shows that \( S_{\textit{GMP}} \) improves the gap between ID and OOD data more effectively.  
Furthermore, the KS test statistics improve ceonsiderably from \( 0.7270, 0.7182, 0.6650, \) and \( 0.6760 \) (\( S_{\textit{MCM}} \)) to \( 0.8724, 0.7798, 0.7219, \) and \( 0.6874 \) (\( S_{\textit{GMP}} \)) across the four OOD datasets. 
These results indicate that \( S_{\textit{GMP}} \) achieves better separation between ID and OOD scores compared to \( S_{\textit{MCM}} \).

\noindent
\textbf{Quantifying modality gap}.
To quantify the modality gap, following \cite{liang2022mind}, we formulate the modality gap as the $\ell_2$ distance between the center of image embeddings and (class) text embeddings, \ie, $\left\|\overrightarrow{\Delta}_{\text{gap}}\right\| = \left\|\frac{1}{C}\sum_{c=1}^{C} P_{\text{txt}, c} - \frac{1}{N_{\text{image}}} \sum_{i=1}^{N_{\text{image}}} I_i \right\|_2$, where $P_{\text{txt}, c}$ is the embedding of $c$-th class, $N_{\text{image}}$ is the number of images, and $I_i$ is the image embedding of the $i$-th image. 
We use the test set of each dataset for calculating the modality gap. 
We also calculate the modality gap between the center of mapped image embeddings $f_\textit{txt-img}(I)$ and image ID prototypes ${P_{\text{img}}}$ as shown in \cref{ablation: tab: modality gap}.
The results demonstrate that \ours significantly reduces the modality gap across all datasets. 
Specifically, compared to the baseline MCM, \ours reduces the original modality gap by approximately 25-30\%, which shows that the design of \ours could effectively reduce the gap. 
Moreover, when comparing text prototypes with our mapped image embeddings $f_{t2i}(I)$, we achieve a dramatic reduction of over 75\% in the modality gap (row 3), indicating that our mapping function effectively bridges the gap between modalities. 
Additionally, the comparison between image ID prototypes and original images (row 4) shows substantial improvement.

\subsection{Details of Benchmarking Datasets}
\label{Appendix: Sec: datasets}

\textbf{ImageNet-100, ImageNet-10, and ImageNet-20}. 
Following MCM~\cite{ming2022delving}, we choose 100/10/20 classes from ImageNet-1k~\cite{krizhevskyImagenetClassificationDeep2012} to form ImageNet-100, ImageNet-10, and ImageNet-20. 
The chosen classes for each dataset are as follows:  
\begin{itemize}
    \item 
    \textbf{ImageNet-100}: n03877845, n03000684, n03110669, n03710721, n02825657, n02113186, n01817953, n04239074, n02002556, n04356056, n03187595, n03355925, n03125729, n02058221, n01580077, n03016953, n02843684, n04371430, n01944390, n03887697, n04037443, n02493793, n01518878, n03840681, n04179913, n01871265, n03866082, n03180011, n01910747, n03388549, n03908714, n01855032, n02134084, n03400231, n04483307, n03721384, n02033041, n01775062, n02808304, n13052670, n01601694, n04136333, n03272562, n03895866, n03995372, n06785654, n02111889, n03447721, n03666591, n04376876, n03929855, n02128757, n02326432, n07614500, n01695060, n02484975, n02105412, n04090263, n03127925, n04550184, n04606251, n02488702, n03404251, n03633091, n02091635, n03457902, n02233338, n02483362, n04461696, n02871525, n01689811, n01498041, n02107312, n01632458, n03394916, n04147183, n04418357, n03218198, n01917289, n02102318, n02088364, n09835506, n02095570, n03982430, n04041544, n04562935, n03933933, n01843065, n02128925, n02480495, n03425413, n03935335, n02971356, n02124075, n07714571, n03133878, n02097130, n02113799, n09399592, n03594945.
    
    \item
    \textbf{ImageNet-10}: n04552348, n04285008, n01530575, n02123597, n02422699, n02107574, n01641577, n03417042, n02389026, n03095699.

    \item
    \textbf{ImageNet-20}: n04147183, n02951358, n02782093, n04389033, n03773504, n02917067, n02317335, n01632458, n01630670, n01631663, n02391049, n01693334, n01697457, n02120079, n02114367, n02132136, n03785016, n04310018, n04266014, n04252077.
    
\end{itemize}

\noindent
\textbf{Other datasets}. 
Similarly, we use subsets from iNaturalist~\cite{van_horn_inaturalist_2018}, SUN~\cite{xiao_sun_2010}, Places~\cite{zhou_places_2018}, and Texture~\cite{cimpoi_describing_2014} as OOD datasets, which are created by Huang and Li~\cite{huang_mos_2021}. 
\begin{itemize}
    \item 
        \textbf{iNaturalist} contains images from the natural world with images from 5089 classes, belonging to 13 super-categories, such as Plantae (Plant), Insecta (Insect), Aves (Bird), Mammalia (Mammal).
        The subset containing 110 plant classes not present in ImageNet-1k is chosen as the OOD test set. 
        The classes are as follows: \textit{Coprosma lucida, Cucurbita foetidissima, Mitella diphylla, Selaginella bigelovii, Toxicodendron vernix, Rumex obtusifolius, Ceratophyllum demersum, Streptopus amplexifolius, Portulaca oleracea, Cynodon dactylon, Agave lechuguilla, Pennantia corymbosa, Sapindus saponaria, Prunus serotina, Chondracanthus exasperatus, Sambucus racemosa, Polypodium vulgare, Rhus integrifolia, Woodwardia areolata, Epifagus virginiana, Rubus idaeus, Croton setiger, Mammillaria dioica, Opuntia littoralis, Cercis canadensis, Psidium guajava, Asclepias exaltata, Linaria purpurea, Ferocactus wislizeni, Briza minor, Arbutus menziesii, Corylus americana, Pleopeltis polypodioides, Myoporum laetum, Persea americana, Avena fatua, Blechnum discolor, Physocarpus capitatus, Ungnadia speciosa, Cercocarpus betuloides, Arisaema dracontium, Juniperus californica, Euphorbia prostrata, Leptopteris hymenophylloides, Arum italicum, Raphanus sativus, Myrsine australis, Lupinus stiversii, Pinus echinata, Geum macrophyllum, Ripogonum scandens, Echinocereus triglochidiatus, Cupressus macrocarpa, Ulmus crassifolia, Phormium tenax, Aptenia cordifolia, Osmunda claytoniana, Datura wrightii, Solanum rostratum, Viola adunca, Toxicodendron diversilobum, Viola sororia, Uropappus lindleyi, Veronica chamaedrys, Adenocaulon bicolor, Clintonia uniflora, Cirsium scariosum, Arum maculatum, Taraxacum officinale officinale, Orthilia secunda, Eryngium yuccifolium, Diodia virginiana, Cuscuta gronovii, Sisyrinchium montanum, Lotus corniculatus, Lamium purpureum, Ranunculus repens, Hirschfeldia incana, Phlox divaricata laphamii, Lilium martagon, Clarkia purpurea, Hibiscus moscheutos, Polanisia dodecandra, Fallugia paradoxa, Oenothera rosea, Proboscidea louisianica, Packera glabella, Impatiens parviflora, Glaucium flavum, Cirsium andersonii, Heliopsis helianthoides, Hesperis matronalis, Callirhoe pedata, Crocosmia × crocosmiiflora, Calochortus albus, Nuttallanthus canadensis, Argemone albiflora, Eriogonum fasciculatum, Pyrrhopappus pauciflorus, Zantedeschia aethiopica, Melilotus officinalis, Peritoma arborea, Sisyrinchium bellum, Lobelia siphilitica, Sorghastrum nutans, Typha domingensis, Rubus laciniatus, Dichelostemma congestum, Chimaphila maculata, Echinocactus texensis.}
    \item 
        \textbf{SUN} contains 899 classes that cover indoor, urban, and natural places. 
        We use the subset that contains 50 natural objects that do not overlap with ImageNet-1k. 
        The classes we use are \textit{badlands, bamboo forest, bayou, botanical garden, canal (natural), canal (urban), catacomb, cavern (indoor), corn field, creek, crevasse, desert (sand), desert (vegetation), field (cultivated), field (wild), fishpond, forest (broadleaf), forest (needleleaf), forest path, forest road, hayfield, ice floe, ice shelf, iceberg, islet, marsh, ocean, orchard, pond, rainforest, rice paddy, river, rock arch, sky, snowfield, swamp, tree farm, trench, vineyard, waterfall (block), waterfall (fan), waterfall (plunge), wave, wheat field, herb garden, putting green, ski slope, topiary garden, vegetable garden, formal garden.}
    \item 
        \textbf{Places} contains photos labeled with scene semantic categories from three macro-classes: Indoor, Nature, and Urban. 
        We use a subset sampled from 50 categories that are not present in ImageNet-1k. 
        The classes we use are 
        \textit{badlands, bamboo forest, canal (natural), canal (urban), corn field, creek, crevasse, desert (sand), desert (vegetation), desert road, field (cultivated), field (wild), field road, forest (broadleaf), forest path, forest road, formal garden, glacier, grotto, hayfield, ice floe, ice shelf, iceberg, igloo, islet, japanese garden, lagoon, lawn, marsh, ocean, orchard, pond, rainforest, rice paddy, river, rock arch, ski slope, sky, snowfield, swamp, swimming hole, topiary garden, tree farm, trench, tundra, underwater (ocean deep), vegetable garden, waterfall, wave, wheat field}.
    \item 
        \textbf{Texture} contains images of textures and abstracted patterns. As no categories overlap with ImageNet1k, we use the entire dataset.
\end{itemize}

\section{Proof of \cref{the: image anchor helps}}
\label{Appendix: Sec: Proof}

We first introduce three necessary assumptions before proceeding with the proof of our theorem.

\begin{assumption}[ID image embeddings follow the distribution centered at ID image prototypes]\label{ass: ID from image prototypes}
Image embeddings $\mathbf{x}_{c}$ are drawn from the symmetric distribution $\mathcal{D}_c$ which is centered at the image prototypes $P_{i,c}$ in the embedding space. 
\end{assumption}

\begin{assumption}[The image-text modality gap exists]\label{ass: modality gap}
The image embeddings are closer to the ground truth class's image prototypes than the text prototypes due to the modality gap.
\end{assumption}

\begin{remark}
    These two assumptions are the basic assumptions about the multi-modal prototypes and the modality gap. 
    Based on the findings from a recent study~\cite{liang2022mind}, these two hold true for most real-world scenarios.
\end{remark}

\begin{assumption}[Models are well trained]\label{ass: well trained}
Cosine similarity between embeddings from the same class distribution $\mathcal{D}_i$ is higher on average than between embeddings from different distributions. 
And the image embeddings from the $c$-th class are equally dissimilar to other classes.
\end{assumption}

\begin{remark}
    Here we only consider the optimal representation extractors without any other assumptions on the models.
\end{remark}

\begin{assumption}[OOD embeddings are not sampled from any ID distribution~\cite{fu_clipscope_2025}]\label{ass: OOD condition}
OOD image embeddings $\mathbf{x}_{\textit{OOD}}$ are drawn from a distribution $\mathcal{D}_{\textit{OOD}}$ different from any $\mathcal{D}_c^{*}, \forall c \in \{1..C\}\text{ and } \forall * \in \{\textit{text}, \textit{image}\}$. 
In addition, the OOD image embeddings are equally similar to all ID text/image classes.
\end{assumption}

\begin{remark}
    This assumption is a general basic assumption used by previous studies~\cite{fu_clipscope_2025} on the OOD data, which provides us with an overall characteristic of the OOD data.
\end{remark}

Under the above assumptions, we first would like to prove the following inequality holds:
\begin{equation}
    \begin{aligned}
        \Delta_{\textit{image}}&=\mathbb{E}\left[\max_{i \in \{1..C\}} \sigma_i^{\textit{image}}(\mathbf{x}_{\textit{ID}}) - \max_{i \in \{1..C\}} \sigma_i^{\textit{image}}(\mathbf{x}_{\textit{OOD}})\right] \\
        \geq  \Delta_{\textit{text}}&=\mathbb{E}\left[\max_{i \in \{1..C\}} \sigma_i^{\textit{text}}(\mathbf{x}_{\textit{ID}}) - \max_{i \in \{1..C\}} \sigma_i^{\textit{text}}(\mathbf{x}_{\textit{OOD}})\right] \\
        & \text{s.t. } \sigma_i^{*} = \frac{\exp(s_{*,i})}{\sum_{k=1}^{N} \exp(s_k)} \text{ and } s_{*,i} = cos(\mathbf{x}, P_{*,i}) \\
        & \ \ \forall * \in \{\textit{image}, \textit{text}\} \,,
    \end{aligned}
\end{equation}
where $\mathbf{x}_{\textit{ID}}$ and $\mathbf{x}_{\textit{OOD}}$ are samples from any ID distributions and OOD samples.

First, for ID samples, with Assumption~\ref{ass: well trained}, we have
\begin{equation}
    \begin{aligned}
        \mathbb{E}\left[\max_{c\in\{1..C\}}\sigma_{c}^{*}(\mathbf{x}_{\textit{ID}})\right] &= \frac{\exp\left(\mu_{\textit{intra}}^{*}\right)}{\exp\left(\mu_{\textit{intra}}^{*}\right) + (C-1) \exp\left(\mu_{\textit{inter}}^{*}\right)} \\
        &\forall * \in \{\textit{text}, \textit{image}\}\,,
    \end{aligned}
\end{equation}
where $\mu_{\textit{intra}}^{*} = \max_{c\in\{1..C\}}\mathbb{E}[\cos(\mathbf{x}_{\textit{ID}}, P_{*,c})]$, $\mu_\textit{inter}^{*} = \mathbb{E}[cos(\mathbf{x}_{\textit{ID}}, I_{c_j})], \forall c_j \in \{1..C\} \setminus \{c_{\textit{ID}}\}$, and $C_{\textit{ID}}$ is the ground-truth class for the input $\mathbf{x}_{\textit{ID}}$.

With assumption~\ref{ass: OOD condition}, for OOD samples, we have
\begin{equation}
    \mathbb{E}\left[\max_{c \in \{1..C\}} \sigma_c^{\textit{image}}(\mathbf{x}_{\textit{OOD}})\right] = \mathbb{E}\left[\max_{c \in \{1..C\}} \sigma_c^{\textit{text}}(\mathbf{x}_{\textit{OOD}})\right] = \frac{1}{C}\,.  
\end{equation}

Then we have
\begin{equation}
    \Delta_{\textit{image}} = \mathbb{E}\left[\max_{c\in\{1..C\}}\sigma_{c}^{\textit{image}}(\mathbf{x}_i)\right] - \frac{1}{C}\,.
\end{equation}

Similarly, we have
\begin{equation}
    \Delta_{\textit{text}} = \mathbb{E}\left[\max_{c\in\{1..C\}}\sigma_{c}^{\textit{text}}(\mathbf{x}_i)\right] - \frac{1}{C}\,.    
\end{equation}

Since $\mu_{\textit{intra}}^{\textit{image}} > \mu_{\textit{intra}}^{\textit{text}}$ (Assumption~\ref{ass: modality gap}), we have
\begin{equation}
    \begin{aligned}
        \Delta_{\textit{image}} & > \Delta_{\textit{text}}  \,.        
    \end{aligned}
\end{equation}

By simple algebra, we have
\begin{equation}
    \begin{aligned}
        &  \mathbb{E}\left[\frac{\max_{i \in \{1..C\}} \sigma_i^{\textit{image}}(\mathbf{x}_{\textit{ID}}) + \max_{i \in \{1..C\}} \sigma_i^{\textit{text}}(\mathbf{x}_{\textit{ID}})}{2} \right.\\
        & \left. - \frac{\max_{i \in \{1..C\}} \sigma_i^{\textit{image}}(\mathbf{x}_{\textit{OOD}}) + \max_{i \in \{1..C\}} \sigma_i^{\textit{text}}(\mathbf{x}_{\textit{OOD}})}{2}\right] \\
        \geq &  \mathbb{E}\left[\max_{i \in \{1..C\}} \sigma_i^{\textit{text}}(\mathbf{x}_{\textit{ID}}) - \max_{i \in \{1..C\}} \sigma_i^{\textit{text}}(\mathbf{x}_{\textit{OOD}})\right]\,, \\
        & \mathbb{E}\left[ S_{\textit{MMP}}(\mathbf{x}_{\textit{ID}}) - S_{\textit{MMP}}(\mathbf{x}_{\textit{OOD}}) \right] \\
        \geq & \mathbb{E}\left[ S_{\textit{MCM}}(\mathbf{x}_{ID}) - S_{\textit{MCM}}(\mathbf{x}_{\textit{OOD}}) \right]\,.
    \end{aligned}
\end{equation}

This completes the proof.

\end{document}